\PassOptionsToPackage{dvipsnames}{xcolor}
\documentclass{article} 
\usepackage{iclr2027_conference,times}

\usepackage{amsmath,amsfonts,bm}

\def\eqref#1{equation~\ref{#1}}

\def\1{\bm{1}}

\DeclareMathAlphabet{\mathsfit}{\encodingdefault}{\sfdefault}{m}{sl}
\SetMathAlphabet{\mathsfit}{bold}{\encodingdefault}{\sfdefault}{bx}{n}

\usepackage{hyperref}
\usepackage{url}

\usepackage{latexsym}
\usepackage{amsfonts}
\usepackage{amsmath}
\usepackage{float}
\usepackage{stfloats} 
\usepackage{booktabs}
\usepackage{xcolor}

\usepackage[T1]{fontenc}
\usepackage[utf8]{inputenc}

\usepackage{microtype}

\usepackage{inconsolata}

\usepackage{graphicx}

\usepackage{cleveref}

\definecolor{TeaserBlue}{HTML}{4788e2}
\definecolor{TeaserOrange}{HTML}{ffaa5a}
\definecolor{MidGuessColor}{HTML}{e38a37}   

\usepackage{xspace}
\definecolor{NoGuessColor}{HTML}{555555}
\colorlet{FinalGuessColor}{TeaserBlue}

\newcommand{\stagelabel}[2]{\textcolor{#1}{#2}\xspace}

\newcommand{\noguess}{\stagelabel{NoGuessColor}{no-guess}}
\newcommand{\Noguess}{\stagelabel{NoGuessColor}{No-guess}}
\newcommand{\midguess}{\stagelabel{MidGuessColor}{mid-guess}}
\newcommand{\Midguess}{\stagelabel{MidGuessColor}{Mid-guess}}
\newcommand{\midguesses}{\stagelabel{MidGuessColor}{mid-guesses}}
\newcommand{\finalguess}{\stagelabel{FinalGuessColor}{final-guess}}
\newcommand{\Finalguess}{\stagelabel{FinalGuessColor}{Final-guess}}
\newcommand{\finalguesses}{\stagelabel{FinalGuessColor}{final-guesses}}
\newcommand{\Finalguesses}{\stagelabel{FinalGuessColor}{Final-guesses}}

\usepackage{enumitem}
\usepackage{subcaption}
\usepackage{multirow}
\usepackage{wrapfig}

\iclrfinalcopy 

\title{Beyond the Commitment Boundary: \\ Probing Epiphenomenal Chain-of-Thought in Large Reasoning Models}

\author{Daniel Scalena$^{1,2,*}$ \qquad
Sara Candussio$^{3,*}$ \qquad
Luca Bortolussi$^3$ \\
\textbf{Elisabetta Fersini}$^2$ \qquad
\textbf{Malvina Nissim}$^1$ \qquad
\textbf{Gabriele Sarti}$^4$ \\
$^1$CLCG, University of Groningen \qquad
$^2$University of Milano-Bicocca \\
$^3$University of Trieste \qquad
$^4$Khoury College of Computer Sciences, Northeastern University \\
$^*$Equal contribution \\
\texttt{d.scalena@rug.nl} \qquad
\texttt{sara.candussio@phd.units.it}
}

\begin{document}
\maketitle
\begin{abstract}
Chain-of-thought (CoT) reasoning is the dominant paradigm for inference-time scaling in language models, yet the causal influence of individual steps on the final answer remains poorly understood. In this work, we use answer logits at the end of each reasoning step to estimate each step's causal importance to the final answer and intermediate guesses, shedding light on the answer formation process of several reasoning model families. Across diverse tasks, we find that reasoning typically crosses a \emph{commitment boundary}, a sharp transition from transient intermediate guesses to a stable, high-confidence answer. This transition often happens in a single step, well before the model's reasoning block ends, and is followed by \emph{epiphenomenal} CoT steps that leave the final answer probability unaltered. Using attention probes, we show that answer-formation stages can be linearly decoded from the activations of intermediate reasoning steps with high accuracy, showing robust generalization to unseen reasoning tasks. We leverage this property for early-exiting reasoning blocks at the commitment boundary location, reducing the length of CoTs up to 55\% with negligible impact on model performance.
\ificlrfinal
\footnote{Code and data: \href{https://github.com/DanielSc4/Commitment-Boundary}{DanielSc4/Commitment-Boundary}}
\else
\footnote{Code and data will be released upon final publication.}
\fi

\end{abstract}

\section{Introduction}
\label{sec:introduction}

Large reasoning models produce Chain-of-Thought (CoT) traces before committing to a final answer~\citep{wei2022chain,openai2024o1,snell2025scaling,guo2025deepseekr1}. While CoT reasoning may seem well-suited for safety monitoring thanks to its legible outputs \citep{korbak-etal-2025-cot-monitorability}, a growing body of work has shown that CoT outputs do not reliably provide faithful accounts of the model's internal computations~\citep{turpin2023language,agarwal2024faithfulnessvsplausibilityunreliability, madsen-etal-2024-self,tutek-etal-2025-measuring}. Notable examples of unfaithful CoT include deliberative language with expressions of uncertainty not reflected in model predictions \citep{venhoff2025understanding} and \emph{performative} steps in which model internals encode the final answer well before its explicit verbalization \citep{boppana2026reasoningtheaterdisentanglingmodel}.

\begin{figure}[t]
    \centering
    \includegraphics[width=\linewidth]{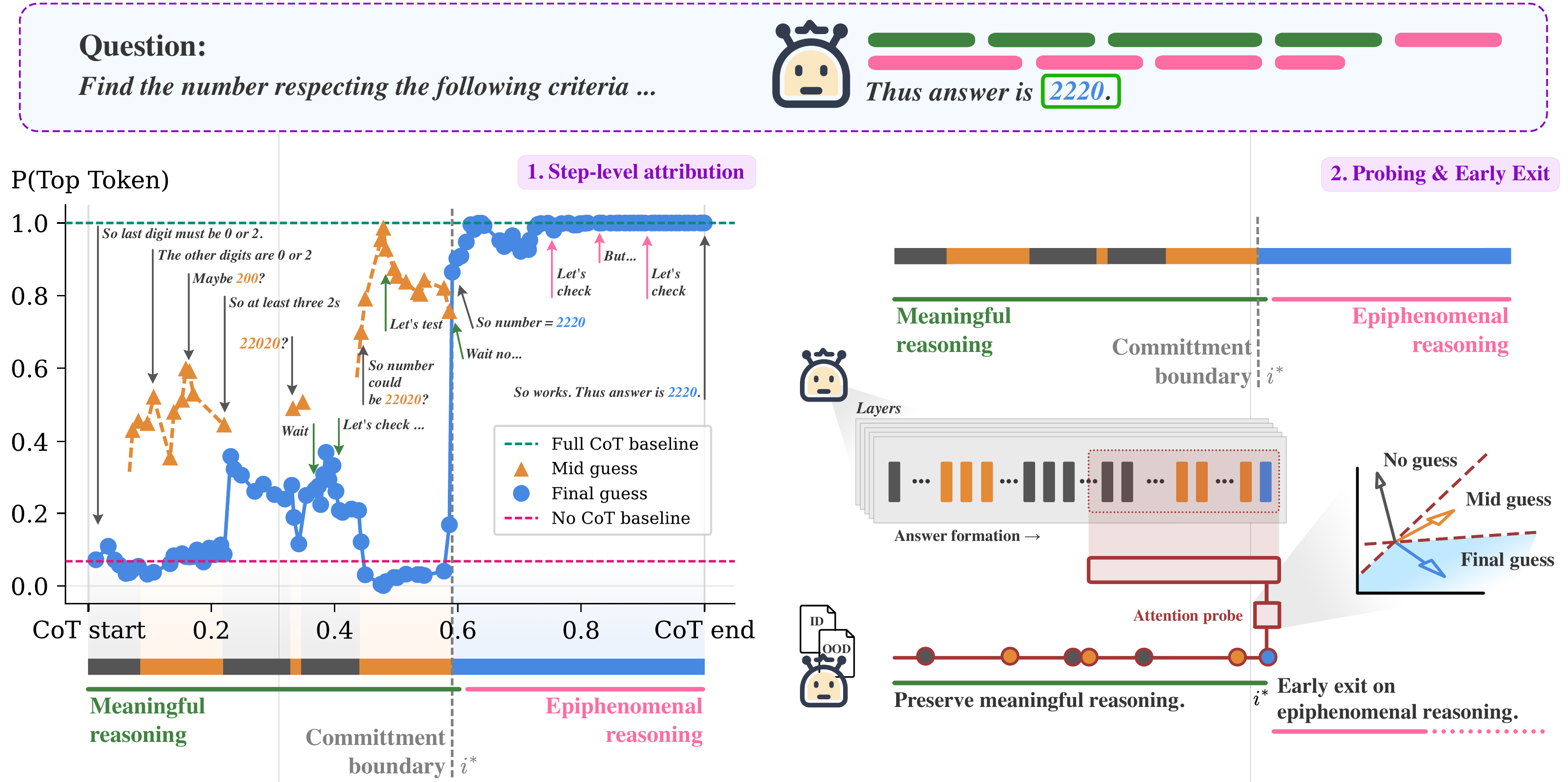}
    \caption{
        Overview of our approach. 
        \textbf{Left:}
        We use early exit to measure the causal contribution of CoT steps to the model's \textcolor{FinalGuessColor}{\textbf{final answer}} and \midguesses probabilities. We frequently encounter a \emph{commitment boundary} $i^*$, marking a sharp transition from meaningful reasoning with \midguesses to a final answer at full-CoT confidence.
        \textbf{Right}: 
        We train lightweight attention probes to predict answer-formation stages from model activations, and use them to early-exit reasoning at $i^*$ to skip performative continuations at little performance cost.
    }
    \label{fig:teaser}
\end{figure}

In this work, we ask: can we track how models converge to their final answer across CoT steps and precisely locate where they stabilize? If so, can we predict when that transition occurs?

We address this via a step-level causal framework that uses CoT early exits to measure shifts in the probability of the model's \emph{final} answer and \midguesses being temporarily entertained at each reasoning step (\Cref{fig:teaser}, top). This yields a causally grounded, fine-grained signal to monitor answer formation throughout reasoning. Across three model families and four popular reasoning tasks, we find that final-answer commitment typically occurs after a single, pivotal reasoning step, producing a large shift in final-answer probabilities. We dub this transition the \textbf{commitment boundary} at step $i^*$, and show that, on average, it falls around the \emph{midpoint} of the CoT. Beyond $i^*$, models often engage in \textbf{epiphenomenal reasoning}: despite frequent hedging and re-verification, the final answer probability remains essentially unchanged. After annotating CoT steps with their respective answer-formation stages (\noguess, \midguess, or \finalguess), we show that lightweight attention probes trained on model activations can reliably predict these stages, with strong performance for in- and out-of-distribution reasoning tasks. Finally, we use these probes to identify the commitment boundary and perform CoT early-exiting at inference time, avoiding superfluous continuations while incurring minimal performance degradation. 
We summarize our contributions as follows:
\begin{itemize}[noitemsep, topsep=2pt]
    \item We introduce a causal framework for step-level analysis of CoT reasoning traces, tracking step-level final commitment and intermediate hypothesis changes in any closed or open-ended verifiable task~(\S\ref{sec:experimental-setup}).
    \item We identify and characterize a commitment boundary $i^*$ across three models and four reasoning tasks~(\S\ref{sec:commitment-boundary}), highlighting the causal effect of pre-boundary reasoning and redundancy of post-boundary steps~(\S\ref{sec:number-perturb}, \S\ref{sec:good-vs-useless}).
    \item We demonstrate that $i^*$ can be reliably located \textbf{in- and out-of-distribution} using lightweight attention probes, suggesting a general structure for answer commitment information across reasoning tasks~(\S\ref{sec:probe}).
    \item We demonstrate that probes enable CoT early-exiting, saving up to 55\% of the reasoning trace with minimal answer accuracy loss~(\S\ref{sec:early-exit}).
\end{itemize}

\section{Related Work}
\label{sec:related-work}

A growing body of work has investigated whether CoT traces faithfully reflect the model's internal reasoning process, and more recently, whether individual steps can be attributed a causal role in producing the final answer.

\paragraph{Faithfulness in CoT reasoning.}
A recurring finding is that CoT explanations provide a false sense of transparency: models asked to explain their reasoning are often inconsistent and unreliable~\citep{madsen-etal-2024-self}, and the generated explanations do not necessarily align with the underlying reasoning process~\citep{agarwal2024faithfulnessvsplausibilityunreliability}.
At a finer grain, \citet{venhoff2025understanding} identifies specific linguistic patterns in CoTs that superficially resemble deliberate reasoning but whose causal status remains unclear.
Other efforts to improve CoT faithfulness through in-context learning, fine-tuning and activation editing have met limited success, suggesting fundamental limitations~\citep{tanneru2024hardnessfaithfulchainofthoughtreasoning,paul-etal-2024-making,shen2026faithcotbench}.
The ambiguity between genuine and epiphenomenal reasoning is a central concern our work addresses directly.

\paragraph{Causal attribution in reasoning traces}
A complementary line of work moves beyond behavioral analysis and attempts to assign causal credit to individual CoT steps. \citet{zhao2026ahamomentsfakeidentifying} identify unfaithful steps by perturbing reasoning steps and observing the downstream effect on the final answer. \citet{lanham2023measuringfaithfulnesschainofthoughtreasoning} evaluate perturbation strategies for measuring CoT faithfulness, finding significant variation across approaches, while \citet{tutek-etal-2025-measuring} use erasure techniques to measure CoT faithfulness for individual reasoning steps. Such perturbation-based methods are inherently noisy: results depend on how the model interprets the intervened CoT, and their causal interpretation is confounded by the model's ability to reason implicitly without explicit CoT~\citep{deng2023implicitchainthoughtreasoning,yee2024faithful}. We sidestep these issues by \emph{truncating} rather than perturbing the trace, and causally estimate the model's confidence in its answers as a function of CoT prefixes of increasing length across reasoning steps.

\paragraph{Interpretability of reasoning}

Recent work has begun decoding the model's internal state to shed light on CoT faithfulness beyond the generated text. 
\citet{wang-etal-2025-chain} exploit mid-layer confidence signals to show that model confidence can be predicted throughout reasoning. \citet{bogdan2026thought} suppress attention between reasoning steps to show that a small subset of sentences has outsized influence on the final answer distribution.

Concurrent to our work,~\citet{boppana2026reasoningtheaterdisentanglingmodel} define performative reasoning as the accuracy gap between a black-box CoT monitor and a probe, or a truncation procedure similar to ours, applied to the same CoT step. Their analysis focuses on backtracking and steps that appear to challenge or revise model guesses, showing that the final answer is decodable from reasoning-step activations before it is explicitly mentioned in the trace. Our notion of \emph{epiphenomenal reasoning} is strictly broader, showing the limited impact of all post-boundary steps on the final answer after the commitment boundary location.
While \citet{boppana2026reasoningtheaterdisentanglingmodel} train probes to predict a limited set of answers from multiple-choice options, we instead focus on predicting whether an answer is temporary or final to identify genuine hypothesis revision across CoT traces. Finally, while their work finds performative reasoning to be difficulty-dependent, we show that model family has a stronger effect on epiphenomenal reasoning.

\paragraph{CoT Efficiency.}
A shared implication of these findings is that not all CoT steps carry equal informational value, motivating work on reducing the computational cost of long reasoning traces via confidence signals~\citep{scalena2026eager,fu2026deep}, compression~\citep{li2026making,xia-etal-2025-tokenskip} or internal geometry~\citep{ge2024model,liu2025kvcachecompressioninference}.
Rather than treating token reduction as the primary goal, we use early exit to confirm the redundancy of post-boundary tokens, thereby grounding inference-time savings in a better understanding of answer-commitment mechanisms.

\section{Measuring CoT Answer Formation}
\label{sec:experimental-setup}

To verify the presence of epiphenomenal reasoning in CoT traces, we introduce a truncation-based causal framework measuring the marginal contribution of each CoT step to the model's final answer. 

\paragraph{Step-Level Attribution.}
Let $P$ be a set of prompt tokens and $C$\footnote{$C$ includes model-specific markers such as \texttt{<think>}. We segment $C$ in sentences at final punctuation marks.} be the resulting CoT. To isolate the effect of the CoT from the answer-generation phase, we construct the answer-forcing sequence $X_{\mathrm{full}} = P + \texttt{[BOT]} + C + \texttt{[EOT]} + S$
where \texttt{[BOT]} and \texttt{[EOT]} are model-dependent beginning- and end-of-thinking markers, $+$ denotes sequence concatenation, and $S$ is a task-specific suffix that elicits a direct answer.\footnote{For mathematical and multiple-choice QA we use the suffix "\texttt{Therefore, the final answer is \textbackslash boxed\{}"}

We segment each decoded CoT $C$ into $n$ sentence-level spans by splitting on terminal punctuation and construct $n+1$ prefix sequences as:
\begin{equation}
    X_i = P + \texttt{[BOT]} + C_i + \texttt{[EOT]} + S \label{eq:xi} \\ 
\end{equation}
where $C_i$ is $C$ truncated at the end of its i-th sentence, $X_0$ is the no-CoT baseline and $X_n = X_{\mathrm{full}}$ is the full-CoT condition. Importantly, we restrict our analysis to traces in which intermediate reasoning improves the output, dropping those in which the model answers correctly when prompted with $X_0$ and those in which the confidence gap between $X_n$ and $X_0$ for the correct answer is $<0.5$.

For each prefix $X_i$, we greedily decode the model's answer $\hat{A}_i$ and compare whether it is semantically equivalent to the full-CoT answer $\hat{A}_n$ using a verifier $\equiv$ (see Appendix~\ref{app:verifier}). Our goal is not to verify the model's correctness, but rather to track the step at which its answer stabilizes to $\hat{A}_n$. This decoupling ensures that ground-truth correctness is not a confounding factor in our assessment.

We define the \emph{step confidence} at CoT step $i$ as:
\begin{equation}
    p_i = \arg \max P( \, \cdot \mid X_i), \quad 0 \le i \le n,
    \label{eq:pi}
\end{equation}
as a scalar proxy for the model's commitment to any answer after reading the first $i$ CoT sentences.
To prevent spurious high confidence caused by token collisions when different answers $\hat{A}_i \not\equiv \hat{A}_n$ share the same leading token, we discard traces containing this collision for at least one prefix $X_i$.\footnote{Collision rates with \finalguesses are, on average, $9\%$, but they vary significantly across examples and tasks. See Appendix~\ref{app:collision} for details.}


\paragraph{Answer formation stages}\label{sec:tau}

We categorize each step $C_i$ as follows, using a pre-defined threshold $\tau$:
\begin{itemize}[noitemsep, topsep=2pt]
    \item \Noguess ($p_i \le \tau$): the model does not commit to any answer.
    \item \Midguess ($p_i > \tau$ and $\hat{A}_i \not\equiv \hat{A}_n$): the model is confident on a \midguess, but has not settled on its final answer.
    \item \Finalguess ($p_i > \tau$ and $\hat{A}_i \equiv \hat{A}_n$): the model's answer matches the full-CoT answer.
\end{itemize}
For our main experiments, we set the threshold as $\tau = p_0 + \tfrac{1}{2}(p_n - p_0)$, i.e., the midpoint between the no-CoT and full-CoT confidences, to reflect the CoT confidence gain in individual traces. Appendix~\ref{app:tau} shows that our conclusions remain robust regardless of the exact value of $\tau$.

\section{Locating the Commitment Boundary}
\label{sec:commitment-boundary}

We evaluate \texttt{gpt-oss-20b}~\citep{openai2025gptoss120bgptoss20bmodel}, \texttt{Qwen3-14B}~\citep{yang2025qwen3technicalreport} and \texttt{gemma-4-26B-A4B-it}~\citep{gemmaModel2026}
across four reasoning datasets: \textbf{MATH-500}~\citep{lightman2023lets}, comprising 500 competition-style mathematics problems, \textbf{AIME~2025}~\citep{aime25}, with harder, more recent mathematical problems, \textbf{ZebraLogic}~\citep{lin2025zebralogic}, with open-ended logical reasoning problems requiring multi-step deductive inference, and \textbf{GPQA-Diamond}~\citep{rein2024gpqa}, with multiple-choice graduate-level questions across biology, chemistry, and physics.

\begin{wrapfigure}{r}{0.5\linewidth}
    \centering
    \vspace{-15pt}
    \includegraphics[width=\linewidth]
        {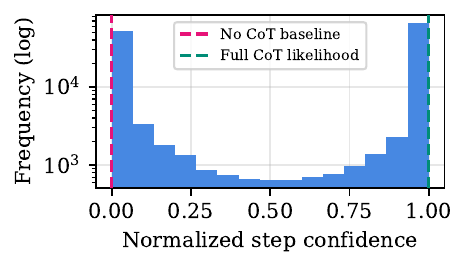}
    \caption{
        \textbf{Answer confidence in reasoning is bimodal.}
        Normalised step confidences $\tilde{p}_i$ across all CoT steps
        in \texttt{gpt-oss-20b} MATH-500 traces. Probability mass
        concentrates near $0$ (no-CoT baseline) and $1$ (full-CoT).
    }
    \label{fig:distrib-p}
    \vspace{-10pt}
\end{wrapfigure}

We begin by observing that \texttt{gpt-oss-20b} normalized step confidences $\tilde{p}_i = (p_i - p_0)/(p_n - p_0)$ across a subset of MATH-500 traces is strongly bimodal (Figure~\ref{fig:distrib-p}): mass concentrates for the most part near $\tilde{p}_i = 0$ (no-CoT level) and $\tilde{p}_i = 1$ (full-CoT level). This suggests that the final answer emerges abruptly as the preferred choice, rather than gradually.

To quantify this phenomenon, we compute the confidence improvement $\Delta_i = p_i - p_{i-1}$ at each step $i$ for the same model, showing that $\max_i \Delta_i$ is on average $4.6$ times larger than the second-largest step improvement (median; 95\% CI: $[4.3, 4.9]$; $n=3{,}119$ traces). These results indicate that in the majority of CoT traces, a single sentence $C_{i^*}$ drives the model from the \noguess regime to its final answer. 

We term this sentence \emph{commitment boundary}, found at position:

\begin{equation}
    i^* = \operatorname*{arg\,max}_{1 \le i \le n} \Delta_i,
    \label{eq:istar}
\end{equation}

\begin{figure}[!t]
    \centering
    \includegraphics[width=1\linewidth]{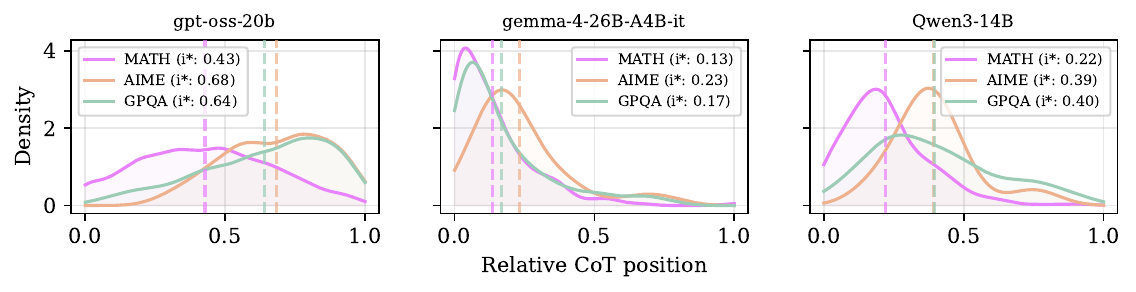}
    \vspace{-24pt}
    \caption{
        Confidence improvement over CoT tokens, across models and datasets. The relative CoT position of the commitment boundary ($i^*/n$) varies greatly across model families, but generally occurs well before the CoT end.
    }
    \label{fig:boundary-position}
\end{figure}

We find that $i^*$ is on average closer to the midpoint of the CoT than to its end, meaning a substantial fraction of CoT tokens are generated after the answer has already stabilized (Figure~\ref{fig:boundary-position}). In particular, we show that \emph{the position of $i^*$ is mainly influenced by model family}, with \texttt{gemma-4-26B-A4B-it} often converging to the final answer after only $13-23\%$ of CoT tokens have been produced, while \texttt{gpt-oss-20b} traces converge to the final solution much later ($43-68\%$). This said, across all models and datasets, we find that \emph{a significant portion of the CoT is produced after the model reaches its final solution}. Finally, we also observe that post-boundary confidence for steps $C_{>i^*}$ remains above the $\tau$ midpoint between CoT and no-CoT solutions across all CoT traces in our evaluation datasets. This indicates that later tokens do not significantly impact the model's final answer.

\subsection{Stress-testing the Commitment Boundary}
\label{sec:number-perturb}

The truncation experiment in \ref{sec:commitment-boundary} shows that the prefix ending at $i^*$ is sufficient for reproducing the full-CoT answer. As an additional test, we ask whether the two sides of the boundary behave differently under targeted perturbations of the reasoning text. We focus in particular on AIME2025 traces, and follow~\citet{zhao2026ahamomentsfakeidentifying} in randomly replacing a fixed percentage of number tokens by adding a small integer offset sampled from $\{-5,\ldots,-1,+1,\ldots,+5\}$. We then append an answer-forcing suffix and greedily decode a new answer, considering the perturbation successful if the resulting answer is not equivalent to the original full-CoT answer $\hat{A}_n$.

We compare two interventions, focusing on the \texttt{gpt-oss-20b} model for this analysis. In the \textsc{post} condition, we perturb only numbers occurring in post-commitment boundary sentences, $C_{>i^*}$, while keeping the full CoT. In the \textsc{pre} condition, we keep only the prefix $X_i^*$ up to the boundary, and perturb only numbers inside it. The unperturbed \textsc{pre} baseline reproduces the full-CoT answer on all retained traces, confirming that $i^*$ is indeed a sufficient decision point before any perturbation is applied.



\begin{figure}[h]
    \centering
    \begin{subfigure}[t]{0.48\linewidth}
        \centering
        \includegraphics[width=\linewidth]
            {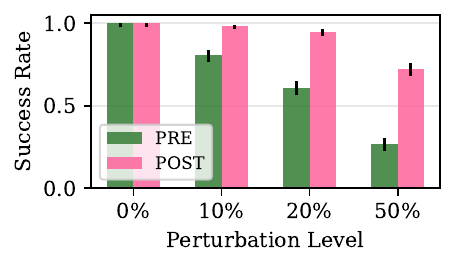}
        \caption{\textbf{Perturbations to $C_{<i^*}$ are most damaging.}
        Fraction of AIME2025 traces whose answer remains
        $\equiv \hat{A}_n$ after numeric corruption.}
        \label{fig:number-perturb}
    \end{subfigure}
    \hfill
    \begin{subfigure}[t]{0.48\linewidth}
        \centering
        \includegraphics[width=\linewidth]
            {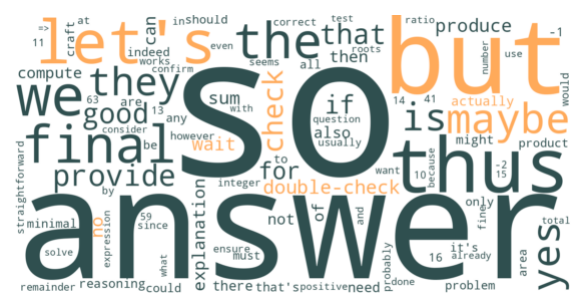}
        \caption{\textbf{Hedging language is frequent after commitment.}
        Content words beginning post-commitment sentences in MATH-500.}
        \label{fig:wordcloud}
    \end{subfigure}
    \caption{\texttt{gpt-oss-20b} reasoning before and after the
    commitment boundary.}
    \label{fig:commitment-analysis}
    \vspace{-8pt}
\end{figure}

Figure~\ref{fig:number-perturb} shows that corrupting the prefix up to $i^*$ is much more destructive than corrupting the post-boundary tail. At $20$\% corruption, \textsc{pre} preserves the answer in only $61$\% of runs, compared to $95$\% for \textsc{post}; at $50$\%, the gap remains large ($27$\% vs.\ $72$\%). These results provide evidence for a phenomenon we term \textbf{epiphenomenal reasoning}: after the model forms its final answer, the remaining tail of reasoning steps has little to no impact on its final prediction. We note this significantly extends the \emph{performative reasoning} phenomenon highlighted by ~\citet{boppana2026reasoningtheaterdisentanglingmodel}, which focused on intermediate reasoning steps encoding the final answer before its explicit mention. In particular, we find that the epiphenomenal tail of reasoning traces often includes steps where the model explicitly challenges or rechecks its own answer, language that superficially resembles genuine deliberation, corroborating previous results by \citet{venhoff2025understanding} (\Cref{fig:wordcloud}).\footnote{See Figure~\ref{fig:inflection-points} and Appendix~\ref{app:post-boundary-examples} for more details.} However, \emph{our truncation experiments confirm these steps have little to no influence on final outputs}.


\subsection{All Meaningful Reasoning Precedes the Commitment Boundary}
\label{sec:good-vs-useless}

\begin{wrapfigure}{r}{0.5\linewidth}
    \centering
    \vspace{-10pt}
    \includegraphics[width=\linewidth]
        {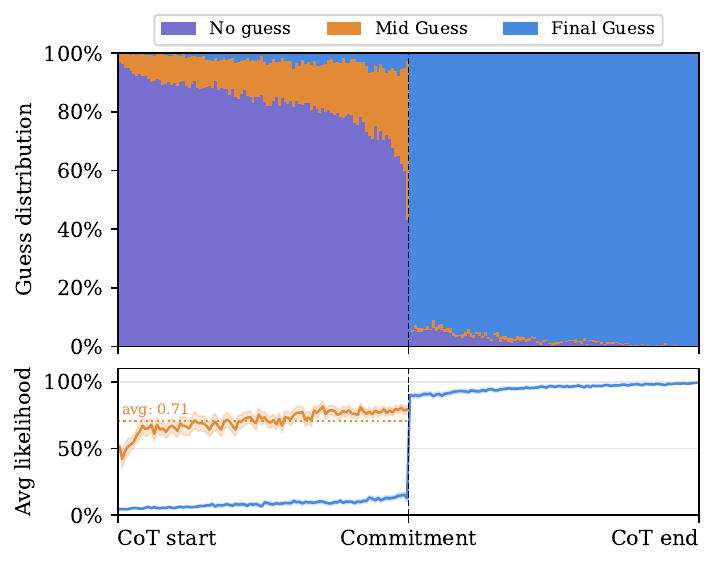}
    \caption{
        \textbf{Pre-boundary steps show genuine answer uncertainty.}
        Distribution of normalised step confidences $\tilde{p}_i$ for
        \texttt{gpt-oss-20b} MATH-500 traces across the three answer
        formation stages. Pre-commitment steps show a growing proportion
        of \midguesses that disappear after the commitment boundary.
    }
    \label{fig:midguess-logprobs}
    \vspace{-10pt}
\end{wrapfigure}
Having established that the commitment boundary $i^*$ marks the start of epiphenomenal reasoning, we now ask whether the pre-boundary CoT is itself informative. \Cref{fig:midguess-logprobs} (top) shows the progression of guesses across \texttt{gpt-oss-20b} MATH-500 traces. Almost all traces begin in the \noguess stage (grey) above our threshold $\tau$, and we observe a large fraction of these transitions directly to the model's \finalguess when reaching the commitment boundary. However, in a good proportion of traces, the model produces one or more \midguesses (orange), each corresponding to a distinct hypothesis the model entertains with high confidence (Avg. likelihood of $0.69$, \Cref{fig:midguess-logprobs}, bottom), and subsequently revises. This provides evidence that the reasoning performed in the pre-commitment region of the CoT is meaningful. In contrast, the guess distribution after the commitment boundary is almost entirely skewed toward the \finalguess, providing additional evidence that post-boundary steps seem to have no causal influence on the model's final answer. We also note that traces containing \midguesses are generally \emph{more than twice as long} as those transitioning directly to the \finalguess ($688$ $\pm 57$ vs.\ $306.15$ $\pm 24$ tokens), with median trace length increasing monotonically with the number of \midguesses on MATH-500, GPQA-Diamond and ZebraLogic (Appendix \ref{app:midguess}). These results expand upon previous findings by \citet{boppana2026reasoningtheaterdisentanglingmodel}: more complex tasks may yield less performative reasoning, since models employ a bigger fraction of their reasoning traces to choose among multiple options.

\paragraph{Different traces of the same problem have consistent \midguesses.} For each question with at least two valid traces, we measure two forms of guess consistency across different traces sampled for the same question: the Jaccard similarity between sets of \midguesses (set overlap), and the standard deviation of the relative position within the CoT at which each \midguess appears.
On MATH-500, \texttt{gpt-oss-20b} achieves a median Jaccard of $0.71$ across trace pairs, indicating that the majority of \midguess values are shared between independently sampled traces of the same problem. The median standard deviation of relative CoT position is $0.005$, roughly half the empirical baseline of $0.011$ obtained by randomly permuting \midguess positions across problems. This suggests that \midguesses tend to recur at similar points across different reasoning chains (Appendix~\ref{app:midguess}).

\section{Probing the Commitment Boundary from Model Activations}
\label{sec:probe}

While useful for post-hoc analysis of reasoning dynamics post-hoc, computing commitment boundary positions $i^*$ for new traces is prohibitively expensive in practical settings, requiring up to $n+1$ forward passes to sample answer completions across steps using our truncation method. However, we hypothesize that the existence of a commitment boundary may reflect a stable internal mechanism rather than a purely behavioral artifact, and that its occurrence could be efficiently decoded from model activations in a single forward pass.

To test this hypothesis, we begin by extracting activations $(\mathbf{h}_1, \dots, \mathbf{h}_n)  \in \mathbb{R}^d$ at a designated layer $\ell$ for the last token of each CoT step $C_1, \dots, C_n$ across all $n$ traces. Each activation $\mathbf{h}_i$ is associated to a label $y_i \in \{\noguess, \midguess, \finalguess\}$ corresponding to the answer formation stage of the corresponding step $C_i$, obtained from our previous truncation experiments. Following \citet{boppana2026reasoningtheaterdisentanglingmodel}, we train a small causal attention probe to perform three-way classification given a fixed context window of $w$ previous CoT steps, enabling its predictive usage at inference time.\footnote{Full probe architecture described in Appendix \ref{app:probe}.}

We train our probes on a stratified split of MATH-500 traces (50\% train / 10\% validation / 10\% test, with 30\% held out for Section~\ref{sec:early-exit}) and use the remaining datasets as test sets spanning various degrees of similarity to the training data distribution, from similar math datasets (AIME 2025) to diverse logic problems (ZebraLogic) and multiple-choice QA (GPQA). We optimize a class-weighted cross-entropy loss using Adam and select the final checkpoint via early stopping on validation macro-$F_1$. At inference time, we predict $\hat{y}_i = \arg\max f(\mathbf{h}_{i - w}, \dots, \mathbf{h}_{i})$.\footnote{Full hyperparameters, architectural choices ($\ell$, $w$), and label distributions are reported in Appendix~\ref{app:probe-sweep}.}

\begin{table*}[t]
\centering
\small
\setlength{\tabcolsep}{5pt}
\begin{tabular}{llcccccc}
\toprule
\textbf{Model} & \textbf{Eval}
    & \textbf{Det.}\,$\uparrow$
    & \textbf{Early-Fire}\,$\downarrow$
    & \textbf{Miss}\,$\downarrow$
    & \textbf{Exact}\,$\uparrow$
    & \textbf{Delay}\,$\downarrow$
    & \textbf{Saved}\,$\uparrow$ \\
\midrule
\multirow{2}{*}{\texttt{gpt-oss-20b}}
  & ID  & 92\%       & 6\%       & 0\%      & 27\%       & 9\%      & 35\% \\
  & OOD & 69 -- 97\% & 3 -- 20\% & 0 -- 0\% & 20 -- 23\% & 3 -- 23\% & 6 -- 35\% \\
\midrule
\multirow{2}{*}{\texttt{gemma-4-26B-A4B-it}}
  & ID  & 62\%       & 16\%       & 25\%       & 15\%       & 30\%       & 55\% \\
  & OOD & 98 -- 99\% & 0 -- 0\% & 1 -- 2\% & 0 -- 7\% & 25 -- 36\% & 40 -- 44\% \\
\midrule
\multirow{2}{*}{\texttt{Qwen3-14B}}
  & ID  & 98\%        & 2\%       & 0\%      & 4\%       & 19\%      & 48\% \\
  & OOD & 78 -- 98\% & 0 -- 21\% & 0 -- 2\% & 4 -- 25\% & 5 -- 47\% & 13 -- 55\% \\
\bottomrule
\end{tabular}
\caption{
    \textbf{Boundary localization diagnostics} for causal attention probes firing at \finalguess positions.
    \textbf{Det.}: \% of positive traces with firing within $k=2$ steps after $i^*$.
    \textbf{Early-Fire}: \% of traces with with firing before $i^*$.
    \textbf{Miss}: \% of positive traces with no firing.
    \textbf{Exact}: \% of traces with firing exactly at $i^*$ ($k{=}0$).
    \textbf{Delay}: Median \# of steps between $i^*$ and probe firing on non-early detections.
    \textbf{Saved}: Median \% of CoT tokens saved across positive traces (misses counted as $0$). ID = MATH-500,
    OOD \textit{min}--\textit{max} ranges span AIME~2025, ZebraLogic, and GPQA-Diamond.
}
\label{tab:probe-diagnostics}
\end{table*}

\paragraph{Boundary localization diagnostics.} Rather than reporting sentence-level classification accuracy, which does not reflect the ability of the probe to precisely detect the location of the commitment boundary, we rely on multiple metrics that summarize the probe's ability to identify the \finalguess location for usage in monitoring and early-exiting:
\begin{itemize}[noitemsep, topsep=2pt]
    \item \textbf{Detection rate (Det.):} Fraction of traces on which the probe fires at or briefly after $i^*$ (using a $k = 2$-sized tolerance window).
    \item \textbf{Early-Fire Rate:} Fraction of traces on which the probe fires \emph{before} $i^*$.
    \item \textbf{Miss Rate:} Fraction of traces for which commitment is never detected.
    \item \textbf{Exact-boundary rate:} Fraction of traces in which the probe fires \emph{exactly at} $i^*$.
    \item \textbf{Detection delay:} Median fraction of traces between the true $i^*$ and the probe nearest post-boundary firing.
    \item \textbf{Saved tokens:} Median fraction of CoT tokens that would be skipped by early-exiting at the probe's predicted location.
\end{itemize}

The tolerance window $k$ is defined as a run of $k$ consecutive \finalguess predictions:

\begin{equation}
i_{\mathrm{exit}}^{(k)}
=
\min\{\, i : \hat y_{i-k+1:i} = {\finalguess}^k \,\},
\label{eq:exit-rule}
\end{equation}

where $\hat{y}_{a:b}$ denotes subsequence $\left( \hat{y}_a, \dots, \hat{y}_b \right)$. The exact-boundary rate metric hence corresponds to a tolerance window of $k=0$.

We find that a tolerance window of $k=2$ provides a good balance between efficiency and precision, providing enough context to elicit a \finalguess prediction without accumulating unnecessary post-boundary tokens.\footnote{See full sweep over $k$ in Appendix~\ref{app:probe-sweep}.} Table~\ref{tab:probe-diagnostics} reports our results across tested models, showing MATH-500 performance as ID and min.-max. performance across the other three datasets as OOD. 

Across most settings, detection rates exceed 90\% and \emph{remain equally high in OOD settings}, while early-fire rates remain low, indicating that the probe fires at the right place or slightly after in the vast majority of traces. The generally low mean detection delay implies a small but non-zero residual tail of post-boundary tokens even when the probe succeeds. The consistent pattern confirms that \emph{the commitment boundary can be reliably detected across model families}.

\begin{figure*}[t]
    \centering
    \includegraphics[width=\linewidth]{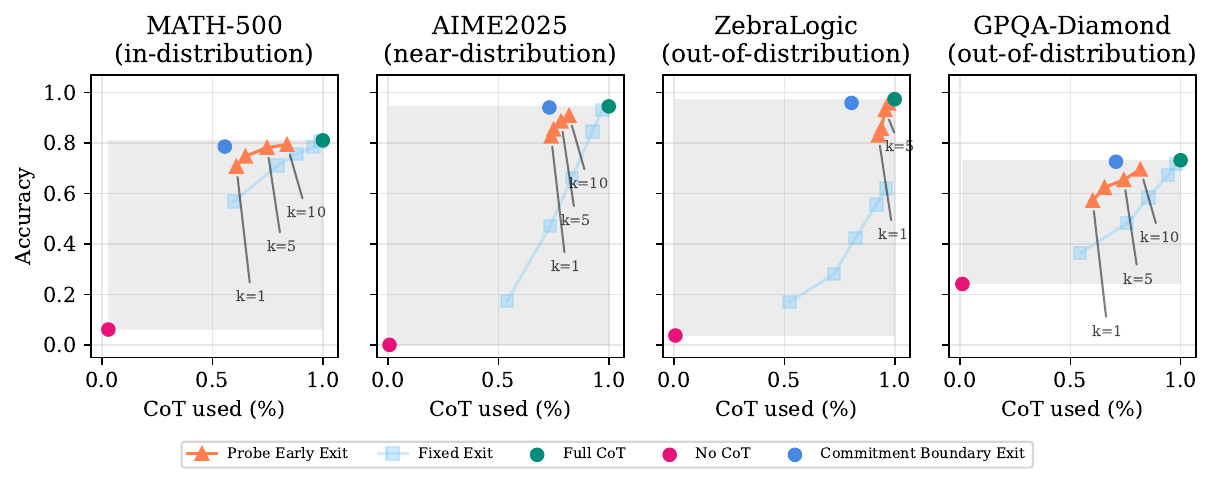}
    \caption{\textbf{Probe-mediated early exit dominates fixed-percentage truncation at every operating point,} with results consistent across in- and out-of-distribution datasets suggesting robust detection capabilities.
}
    \label{fig:early-exit}
\end{figure*}

\section{Early Exiting at the Commitment Boundary}
\label{sec:early-exit}
\Cref{sec:probe} probes predict a per-step answer class $\hat{y}_i$ that can be computed during generation. If the model is genuinely aware of its commitment boundary, then halting generation as soon as $\hat{y}_i = \finalguess$ should preserve accuracy while skipping the post-boundary tokens that Section~\ref{sec:commitment-boundary} identified as output-redundant.

To test the early-exiting performance of trained probes, we adopt the same $k=2$ tolerance window used in \Cref{sec:probe}. If no exit index exists, we fall back to the full CoT. At inference time, we evaluate the effectiveness of our approach by truncating the trace at the end of sentence $i_{\mathrm{exit}}^{(k)}$, appending the model-specific end-of-thinking marker and the task-dependent answer-forcing suffix $S$, and decoding the answer greedily as in our truncation experiments of \Cref{sec:commitment-boundary} (see Appendix~\ref{app:verifier}).

We compare against three black-box baselines for early-exiting:

\begin{itemize}[noitemsep, topsep=2pt]
    \item \textbf{Full CoT}: Use the entire generated trace as an upper bound on accuracy, with no token saving.
    \item \textbf{No CoT}: Baseline performance without access to reasoning tokens.
    \item \textbf{Fixed Exit}: Using the insight that commitment boundary is far from the true end of the CoT (Figure~\ref{fig:boundary-position}), we truncate each trace at a fixed fraction $\rho \in \{50, 70, 80, 90, 95\}\%$ of its total CoT length, using the same answer-forcing protocol.
\end{itemize}

The Fixed Exit baseline acts as a non-adaptive baseline for our evaluation, showing whether the probe's exit predictions can outperform average statistics for the commitment boundary location reported in~\Cref{sec:commitment-boundary}.\footnote{We test whether the probe recognizes valid early-exit locations, and do not aim for an optimal early-exit method. A thorough comparison with established early-exit baselines is hence beyond the scope of this work.}

For each variant we report (i) accuracy relative to the ground-truth answer $a^*$, (ii) the average fraction of CoT tokens saved $1 - n_{\mathrm{kept}}/n_{\mathrm{total}}$, and (iii) the boundary localisation error $\mathbb{E}[(i_{\mathrm{exit}}^{(k)} - i^*)/n]$, signed so that negative values indicate exiting before the true commitment boundary.

Figure~\ref{fig:early-exit} summarises the trade-off curve across our tested datasets.\footnote{See~\Cref{fig:combined_accuracy_vs_cot} for results on other models.} The probe consistently outperforms fixed-percentage truncation at every operating point, confirming that it exploits per-trace structure to improve performance adaptively.
On the in-distribution MATH-500 dataset, the fixed-percentage baseline drops from 80\% full-CoT accuracy to 57\%, whereas the probe retains most of the performance across all values of $k$. At $k{=}1$, it recovers 89\% of full-CoT accuracy while saving 39\% of CoT tokens on average, with a median boundary error of just one sentence past $i^*$. Increasing $k$ trades token savings for accuracy monotonically: $k{=}5$ reaches 98\% of full-CoT accuracy at 26\% token savings.

On AIME~2025, ZebraLogic, and GPQA-Diamond, the MATH-500 trained probe continues to consistently outperform fixed baselines with small accuracy loss compared to full-CoT (at most 11\% on ZebraLogic, 97\% $\rightarrow$ 86\%) while saving 25\% / 6\% / 35\% of CoT tokens respectively. The Fixed Exit controls lose more accuracy on every OOD dataset, with the largest gap for AIME~2025, where longer traces make early exiting most beneficial. Together with probe results of Section~\ref{sec:probe}, these findings suggest that \emph{the commitment boundary is a stable, model-internal property that can be linearly decoded and reliably used for early exiting at inference time}.

\section{Conclusion}
\label{sec:conclusion}

We introduced a step-level causal attribution framework based on CoT early exits to study the answer formation process through the lens of large reasoning models' internal processing. Across three model families and four reasoning benchmarks, we identified the \emph{commitment boundary}: a sharp, often single-step transition after which the model's final answer probability is essentially fixed. Through truncation and numeric perturbation experiments, we showed that pre-boundary steps causally shape the answer, with \midguesses frequently arising with moderate confidence and later getting revised. On the contrary, we show that post-boundary steps form an \emph{epiphenomenal} tail of hedging and re-verification where the final answer confidence remains high and mostly unchanged. Its relative position depends primarily on the model family rather than task difficulty, with up to 87\% of generated reasoning tokens occurring after the answer has stabilized in some configurations.
We showed that answer-formation stages are linearly decodable from intermediate activations using lightweight causal attention probes that generalize across in- and out-of-distribution reasoning tasks. Used as online exit signals, these probes consistently outperform fixed-percentage truncation, saving up to 35\% of CoT tokens with minimal accuracy degradation. Together, these results suggest that the commitment boundary is not a behavioral artifact but a stable signal of answer commitment in reasoning models.

Our findings have important implications for inference-time efficiency and reasoning faithfulness. In particular, the frequent occurrence of epiphenomenal reasoning suggests current models already engage in unfaithful reasoning, and reinforces the need for monitoring tools that combine surface-level inspection with richer signals extracted from model internals. We hope our framework provides a useful lens for future work designing training objectives that align the externalized trace more tightly with the underlying answer-formation process.

\section*{Limitations}
\label{sec:limitations}

\paragraph{Greedy decoding and answer elicitation.}
Our step-level attribution relies on greedy decoding from each truncated prefix $X_i$ with a task-specific answer-forcing suffix $S$. This choice yields a deterministic, reproducible signal but ignores the full predictive distribution: a step could meaningfully shift probability mass between answers without changing the arg-max, and would be invisible to our framework. Relatedly, the suffix $S$ is itself a soft intervention that pushes the model to commit to an answer that it might have continued to deliberate over. Our conclusions, therefore, concern the answer the model would commit to if forced at step $i$, which is a useful but not exhaustive notion of commitment. Sampling-based or distribution-level extensions of the framework are a natural direction for future work.

\paragraph{Trace and token filtering.}
We restrict our analysis to traces where the no-CoT baseline is incorrect and discard traces with first-token collisions between intermediate and final answers (Appendix~\ref{app:collision}). While necessary to avoid confounds, these filters bias our sample toward problems where CoT is genuinely informative and where answers are tokenization-distinguishable.

\paragraph{Sentence-level granularity.}
We segment CoTs at terminal punctuation, treating sentences as the atomic unit of reasoning. This is a coarse approximation: commitment may occur mid-sentence, and our reported boundary positions $i^*/n$ should be read as upper bounds on localization precision. Finer-grained attribution (clause- or token-level) would likely shift $i^*$ earlier in some traces, but was too computationally prohibitive for our approach.

\paragraph{Model and task coverage.}
Our experiments span three open-weight model families (\texttt{gpt-oss-20b}, \texttt{Qwen3-14B}, \texttt{gemma-4-26B-A4B-it}) and four reasoning benchmarks dominated by mathematics, logic, and multiple-choice science. We do not evaluate on agentic, tool-using, or long-horizon generation tasks, where the notion of a single final answer is less well-defined and the commitment boundary may not exist or may take a qualitatively different form. We also do not study closed-weight frontier models, where activation-based probing is not possible.

\paragraph{Probe supervision and labels.}
Our probes are trained on labels derived from the same truncation procedure they are meant to replace, making them prone to inherit the biases of our methodology. While they seem to generalize to OOD benchmarks, performance on substantially different domains (code, open-ended QA, multilingual reasoning) remains a promising avenue for future work.

\paragraph{Early-exit safety.}
Probe-mediated early exit trades a small accuracy loss for substantial token savings, but the loss is not zero, and the early-fire rate is non-negligible on some OOD settings (up to 21\% on \texttt{Qwen3-14B}). In deployment contexts where individual answers carry high stakes (medical, legal, safety-critical reasoning), even a small fraction of prematurely truncated traces may be unacceptable. Calibrating exit thresholds per task and per acceptable-risk profile is left to future work.

\paragraph{Interpretation of epiphenomenal reasoning.}
Finally, we caution against over-interpreting epiphenomenal reasoning as evidence that post-boundary tokens are fully functionally useless to the model. Our framework measures their effect on the \emph{elicited final answer} under answer-forcing; it does not rule out roles such as calibrating internal confidence, supporting follow-up turns, or shaping behavior under different decoding strategies. The post-boundary tail may also serve training-time objectives (e.g., RL reward shaping) that are not visible in a single inference pass. Characterizing what epiphenomenal reasoning is for remains a promising open question.

\ificlrfinal
\section*{Acknowledgements}
The work of Daniel Scalena has been partially funded by MUR under the grant ReGAInS, \textit{Dipartimenti di Eccellenza 2023-2027} of the Department of Informatics, Systems and Communication at the University of Milano-Bicocca.
The work of Sara Candussio has been funded by Fondo Sociale Europeo Plus of Regione Autonoma Friuli Venezia Giulia. 
The work of Elisabetta Fersini has been also partially funded by the European Union -- NextGenerationEU under the National Research Centre For HPC, Big Data and Quantum Computing - Spoke 9 - Digital Society and Smart Cities (PNRR-MUR).
Gabriele Sarti is supported by the National Deep Inference Fabric (NDIF) project (U.S. NSF Award IIS-2408455).
We also thank the Center for Information Technology of the University of Groningen for providing access to the H\'abr\'ok high-performance computing cluster used for part of the experiments.
\fi

\clearpage

\bibliography{custom,anthology-1,anthology-2}

@misc{boppana2026reasoningtheaterdisentanglingmodel,
    title={Reasoning Theater: Disentangling Model Beliefs from Chain-of-Thought}, 
    author={Siddharth Boppana and Annabel Ma and Max Loeffler and Raphael Sarfati and Eric Bigelow and Atticus Geiger and Owen Lewis and Jack Merullo},
    year={2026},
    eprint={2603.05488},
    archivePrefix={arXiv},
    primaryClass={cs.CL},
    url={https://arxiv.org/abs/2603.05488}, 
}

@misc{zhao2026ahamomentsfakeidentifying,
    title={Can Aha Moments Be Fake? Identifying True and Decorative Thinking Steps in Chain-of-Thought}, 
    author={Jiachen Zhao and Yiyou Sun and Weiyan Shi and Dawn Song},
    year={2026},
    eprint={2510.24941},
    archivePrefix={arXiv},
    primaryClass={cs.LG},
    url={https://arxiv.org/abs/2510.24941}, 
}

@misc{kydlicek2024mathverify,
    author = {Hynek Kydlíček},
    title= {Math-Verify: Math Verification Library},
    year = {2024},
    howpublished = {\url{https://github.com/huggingface/math-verify}},
    note = {Version 0.6.1}
}

@inproceedings{scalena2026eager,
    title={EAGER: Entropy-Aware GEneRation for Adaptive Inference-Time Scaling},
    author={Scalena, Daniel and Zotos, Leonidas and Fersini, Elisabetta and Nissim, Malvina and {\"U}st{\"u}n, Ahmet},
    booktitle={Proceedings of the 43rd International Conference on Machine Learning (ICML 2026)},
    year={2026},
    url={https://arxiv.org/abs/2510.11170}
}

@inproceedings{venhoff2025understanding,
    title={Understanding Reasoning in Thinking Language Models via Steering Vectors},
    author={Constantin Venhoff and Iv{\'a}n Arcuschin and Philip Torr and Arthur Conmy and Neel Nanda},
    booktitle={Workshop on Reasoning and Planning for Large Language Models},
    year={2025},
    url={https://openreview.net/forum?id=OwhVWNOBcz}
}

@misc{agarwal2024faithfulnessvsplausibilityunreliability,
    title={Faithfulness vs. Plausibility: On the (Un)Reliability of Explanations from Large Language Models}, 
    author={Chirag Agarwal and Sree Harsha Tanneru and Himabindu Lakkaraju},
    year={2024},
    eprint={2402.04614},
    archivePrefix={arXiv},
    primaryClass={cs.CL},
    url={https://arxiv.org/abs/2402.04614}, 
}

@misc{lanham2023measuringfaithfulnesschainofthoughtreasoning,
      title={Measuring Faithfulness in Chain-of-Thought Reasoning}, 
      author={Tamera Lanham and Anna Chen and Ansh Radhakrishnan and Benoit Steiner and Carson Denison and Danny Hernandez and Dustin Li and Esin Durmus and Evan Hubinger and Jackson Kernion and Kamilė Lukošiūtė and Karina Nguyen and Newton Cheng and Nicholas Joseph and Nicholas Schiefer and Oliver Rausch and Robin Larson and Sam McCandlish and Sandipan Kundu and Saurav Kadavath and Shannon Yang and Thomas Henighan and Timothy Maxwell and Timothy Telleen-Lawton and Tristan Hume and Zac Hatfield-Dodds and Jared Kaplan and Jan Brauner and Samuel R. Bowman and Ethan Perez},
      year={2023},
      eprint={2307.13702},
      archivePrefix={arXiv},
      primaryClass={cs.AI},
      url={https://arxiv.org/abs/2307.13702}, 
}

@misc{tanneru2024hardnessfaithfulchainofthoughtreasoning,
      title={On the Hardness of Faithful Chain-of-Thought Reasoning in Large Language Models}, 
      author={Sree Harsha Tanneru and Dan Ley and Chirag Agarwal and Himabindu Lakkaraju},
      year={2024},
      eprint={2406.10625},
      archivePrefix={arXiv},
      primaryClass={cs.CL},
      url={https://arxiv.org/abs/2406.10625}, 
}

@misc{deng2023implicitchainthoughtreasoning,
      title={Implicit Chain of Thought Reasoning via Knowledge Distillation}, 
      author={Yuntian Deng and Kiran Prasad and Roland Fernandez and Paul Smolensky and Vishrav Chaudhary and Stuart Shieber},
      year={2023},
      eprint={2311.01460},
      archivePrefix={arXiv},
      primaryClass={cs.CL},
      url={https://arxiv.org/abs/2311.01460}, 
}

@inproceedings{yee2024faithful,
    title={Faithful and Unfaithful Error Recovery in Chain of Thought},
    author={Evelyn Yee and Alice Li and Chenyu Tang and Yeon Ho Jung and Ramamohan Paturi and Leon Bergen},
    booktitle={First Conference on Language Modeling},
    year={2024},
    url={https://openreview.net/forum?id=IPZ28ZqD4I}
}

@inproceedings{fu2026deep,
title={Deep Think with Confidence},
author={Yichao Fu and Xuewei Wang and Hao Zhang and Yuandong Tian and Jiawei Zhao},
booktitle={The Fourteenth International Conference on Learning Representations},
year={2026},
url={https://openreview.net/forum?id=8LqHs0KIM7}
}

@misc{lightman2023lets,
   title={Let's Verify Step by Step}, 
   author={Hunter Lightman and Vineet Kosaraju and Yura Burda and Harri Edwards and Bowen Baker and Teddy Lee and Jan Leike and John Schulman and Ilya Sutskever and Karl Cobbe},
   year={2023},
   eprint={2305.20050},
   archivePrefix={arXiv},
   primaryClass={cs.LG}
}

@misc{aime25,
    title={American Invitational Mathematics Examination ({AIME}) 2025}, 
    author={Zhang, Yifan and Math-AI, Team},
    year={2025},
}

@inproceedings{lin2025zebralogic,
    title={ZebraLogic: On the Scaling Limits of {LLM}s for Logical Reasoning},
    author={Bill Yuchen Lin and Ronan Le Bras and Kyle Richardson and Ashish Sabharwal and Radha Poovendran and Peter Clark and Yejin Choi},
    booktitle={Forty-second International Conference on Machine Learning},
    year={2025},
    url={https://openreview.net/forum?id=sTAJ9QyA6l}
}

@inproceedings{rein2024gpqa,
    title={{GPQA}: A Graduate-Level Google-Proof Q\&A Benchmark},
    author={David Rein and Betty Li Hou and Asa Cooper Stickland and Jackson Petty and Richard Yuanzhe Pang and Julien Dirani and Julian Michael and Samuel R. Bowman},
    booktitle={First Conference on Language Modeling},
    year={2024},
    url={https://openreview.net/forum?id=Ti67584b98}
}

@misc{openai2025gptoss120bgptoss20bmodel,
    title={gpt-oss-120b \& gpt-oss-20b Model Card},
    author={OpenAI},
    year={2025},
    eprint={2508.10925},
    archivePrefix={arXiv},
    primaryClass={cs.CL},
    url={https://arxiv.org/abs/2508.10925}
}

@misc{gemmaModel2026,
    title={Gemma 4: Byte for byte, the most capable open models},
    author={Google},
    year={2026},
    url={https://blog.google/innovation-and-ai/technology/developers-tools/gemma-4/}
}

@misc{yang2025qwen3technicalreport,
      title={Qwen3 Technical Report}, 
      author={An Yang and Anfeng Li and Baosong Yang and Beichen Zhang and Binyuan Hui and Bo Zheng and Bowen Yu and Chang Gao and Chengen Huang and Chenxu Lv and Chujie Zheng and Dayiheng Liu and Fan Zhou and Fei Huang and Feng Hu and Hao Ge and Haoran Wei and Huan Lin and Jialong Tang and Jian Yang and Jianhong Tu and Jianwei Zhang and Jianxin Yang and Jiaxi Yang and Jing Zhou and Jingren Zhou and Junyang Lin and Kai Dang and Keqin Bao and Kexin Yang and Le Yu and Lianghao Deng and Mei Li and Mingfeng Xue and Mingze Li and Pei Zhang and Peng Wang and Qin Zhu and Rui Men and Ruize Gao and Shixuan Liu and Shuang Luo and Tianhao Li and Tianyi Tang and Wenbiao Yin and Xingzhang Ren and Xinyu Wang and Xinyu Zhang and Xuancheng Ren and Yang Fan and Yang Su and Yichang Zhang and Yinger Zhang and Yu Wan and Yuqiong Liu and Zekun Wang and Zeyu Cui and Zhenru Zhang and Zhipeng Zhou and Zihan Qiu},
      year={2025},
      eprint={2505.09388},
      archivePrefix={arXiv},
      primaryClass={cs.CL},
      url={https://arxiv.org/abs/2505.09388}, 
}

@inproceedings{wei2022chain,
author = {Wei, Jason and Wang, Xuezhi and Schuurmans, Dale and Bosma, Maarten and Ichter, Brian and Xia, Fei and Chi, Ed H. and Le, Quoc V. and Zhou, Denny},
title = {Chain-of-thought prompting elicits reasoning in large language models},
year = {2022},
isbn = {9781713871088},
publisher = {Curran Associates Inc.},
address = {Red Hook, NY, USA},
booktitle = {Proceedings of the 36th International Conference on Neural Information Processing Systems},
articleno = {1800},
numpages = {14},
location = {New Orleans, LA, USA},
series = {NIPS '22}
}

@inproceedings{
snell2025scaling,
title={Scaling {LLM} Test-Time Compute Optimally Can be More Effective than Scaling Parameters for Reasoning},
author={Charlie Victor Snell and Jaehoon Lee and Kelvin Xu and Aviral Kumar},
booktitle={The Thirteenth International Conference on Learning Representations},
year={2025},
url={https://openreview.net/forum?id=4FWAwZtd2n}
}

@article{openai2024o1,
  title={OpenAI o1 System Card},
  author={{OpenAI}},
  journal={arXiv preprint arXiv:2412.16720},
  year={2024},
  url={https://arxiv.org/abs/2412.16720}
}

@article{guo2025deepseekr1,
  title   = {{DeepSeek-R1} incentivizes reasoning in {LLMs} through reinforcement learning},
  author  = {Guo, Daya and Yang, Dejian and Zhang, Haowei and Song, Junxiao and Wang, Peiyi and Zhu, Qihao and Xu, Runxin and Zhang, Ruoyu and Ma, Shirong and Bi, Xiao and others},
  journal = {Nature},
  volume  = {645},
  number  = {8081},
  pages   = {633--638},
  year    = {2025},
  month   = sep,
  publisher = {Nature Publishing Group},
  doi     = {10.1038/s41586-025-09422-z},
  url     = {https://www.nature.com/articles/s41586-025-09422-z},
}

@inproceedings{
turpin2023language,
title={Language Models Don't Always Say What They Think: Unfaithful Explanations in Chain-of-Thought Prompting},
author={Miles Turpin and Julian Michael and Ethan Perez and Samuel R. Bowman},
booktitle={Thirty-seventh Conference on Neural Information Processing Systems},
year={2023},
url={https://openreview.net/forum?id=bzs4uPLXvi}
}

@inproceedings{
shen2026faithcotbench,
title={FaithCoT-Bench: Benchmarking Instance-Level Faithfulness of Chain-of-Thought Reasoning},
author={Xu Shen and Song Wang and Zhen Tan and Laura Yao and Xinyu Zhao and Kaidi Xu and Xin Wang and Tianlong Chen},
booktitle={The Fourteenth International Conference on Learning Representations},
year={2026},
url={https://openreview.net/forum?id=lN3yKqqzF1}
}

@misc{korbak-etal-2025-cot-monitorability,
      title={Chain of Thought Monitorability: A New and Fragile Opportunity for AI Safety}, 
      author={Tomek Korbak and Mikita Balesni and Elizabeth Barnes and Yoshua Bengio and Joe Benton and Joseph Bloom and Mark Chen and Alan Cooney and Allan Dafoe and Anca Dragan and Scott Emmons and Owain Evans and David Farhi and Ryan Greenblatt and Dan Hendrycks and Marius Hobbhahn and Evan Hubinger and Geoffrey Irving and Erik Jenner and Daniel Kokotajlo and Victoria Krakovna and Shane Legg and David Lindner and David Luan and Aleksander Mądry and Julian Michael and Neel Nanda and Dave Orr and Jakub Pachocki and Ethan Perez and Mary Phuong and Fabien Roger and Joshua Saxe and Buck Shlegeris and Martín Soto and Eric Steinberger and Jasmine Wang and Wojciech Zaremba and Bowen Baker and Rohin Shah and Vlad Mikulik},
      year={2025},
      eprint={2507.11473},
      archivePrefix={arXiv},
      primaryClass={cs.AI},
      url={https://arxiv.org/abs/2507.11473}, 
}

@misc{bogdan2026thought,
    title={Thought Anchors: Which {LLM} Reasoning Steps Matter?},
    author={Paul C. Bogdan and Uzay Macar and Neel Nanda and Arthur Conmy},
    year={2026},
    url={https://openreview.net/forum?id=6NUtPO9PdV}
}

@inproceedings{li2026making,
    title={Making Slow Thinking Faster: Compressing {LLM} Chain-of-Thought via Step Entropy},
    author={Zeju Li and Jianyuan Zhong and Ziyang Zheng and Xiangyu Wen and Zhijian Xu and Yingying Cheng and Fan Zhang and Qiang Xu},
    booktitle={The Fourteenth International Conference on Learning Representations},
    year={2026},
    url={https://openreview.net/forum?id=cGLqQfS5wH}
}

@inproceedings{ge2024model,
    title={Model Tells You What to Discard: Adaptive {KV} Cache Compression for {LLM}s},
    author={Suyu Ge and Yunan Zhang and Liyuan Liu and Minjia Zhang and Jiawei Han and Jianfeng Gao},
    booktitle={The Twelfth International Conference on Learning Representations},
    year={2024},
    url={https://openreview.net/forum?id=uNrFpDPMyo}
}

@misc{liu2025kvcachecompressioninference,
      title={KV Cache Compression for Inference Efficiency in LLMs: A Review}, 
      author={Yanyu Liu and Jingying Fu and Sixiang Liu and Yitian Zou and You Fu and Jiehan Zhou and Shouhua Zhang},
      year={2025},
      eprint={2508.06297},
      archivePrefix={arXiv},
      primaryClass={cs.DC},
      url={https://arxiv.org/abs/2508.06297}, 
}

@inproceedings{fiottokaufman2024nnsightndifdemocratizingaccess,
 author = {Fiotto-Kaufman, Jaden and Loftus, Alexander and Todd, Eric and Brinkmann, Jannik and Pal, Koyena and Troitskii, Dmitrii and Ripa, Michael and Belfki, Adam and Rager, Can and Juang, Caden and Mueller, Aaron and Marks, Samuel and Sen Sharma, Arnab and Lucchetti, Francesca and Prakash, Nikhil and Brodley, Carla and Guha, Arjun and Bell, Jonathan and Wallace, Byron and Bau, David},
 booktitle = {International Conference on Learning Representations},
 editor = {Y. Yue and A. Garg and N. Peng and F. Sha and R. Yu},
 pages = {92337--92370},
 title = {NNsight and NDIF: Democratizing Access to Open-Weight Foundation Model Internals},
 url = {https://proceedings.iclr.cc/paper_files/paper/2025/file/e6c65eb9b56719c1aa45ff73874de317-Paper-Conference.pdf},
 volume = {2025},
 year = {2025}
}
\bibliographystyle{iclr2027_conference}

\clearpage

\appendix

\section{Candidate solutions verifier}\label{app:verifier}
For each prefix $X_i$, we greedily decode an intermediate answer $\hat{A}_i$ (continuation capped at 64 tokens) and compare it to the full-chain answer $\hat{A}_n$. This comparison requires a semantic equivalence relation $\equiv$ robust to surface-form variation, therefore avoiding to introduce spurious mismatches into the attribution signal.

For consistency across mathematical and multiple-choice benchmarks, all non-code datasets use the same boxed-answer suffix, prompting the model to emit its final answer inside a \verb|\boxed{}| environment. For multiple-choice tasks, the verifier subsequently extracts and normalizes the boxed option letter.

Answer strings are however normalized by stripping surrounding whitespace and math-mode delimiters, removing enclosing \verb|\text{...}| wrappers, and collapsing internal whitespace (e.g. \texttt{p-q} vs.\texttt{p - q}, MATH-500, Q1). If the normalized reference and candidate strings match exactly, they are
immediately declared equivalent.

Because our evaluation benchmarks span heterogeneous answer formats, the verifier dispatches automatically to domain-specific comparison branches to spot non-trivial equivalent guesses.

\paragraph{For mathematical benchmarks} we use the \texttt{math-verify} library
\citep{kydlicek2024mathverify} which extracts mathematical expressions, parses them symbolically via \texttt{latex2sympy2\_extended}, and verifies equivalence using SymPy. To reduce formatting-related parsing failures, we additionally normalize \verb|\tfrac| and \verb|\dfrac| to \verb|\frac| and strip \verb|\left| and \verb|\right| delimiters.
For each answer we evaluate multiple surface variants (with and without math-mode delimiters, before and after normalization), and declare two answers equivalent if any variant pair is successfully verified. This procedure is robust to common algebraic simplifications (\verb|\frac{3}{2}}| vs. $1.5$, e.g. MATH-500 q20), symbolic rewritings, and minor \LaTeX formatting differences (\verb|\tfrac{1}{2}| vs.\ \verb|\frac{1}{2}|, e.g. MATH-500 q0).

\paragraph{For multiple-choice benchmarks} (e.g.\ GPQA-Diamond), we extract the predicted option letter using a lightweight regular-expression matcher after normalization. This canonicalizes variants such as \texttt{A}, \texttt{(A)}, and \texttt{A.}. If option extraction fails, comparison falls back to normalized string matching. Two answers are considered equivalent if and only if their canonicalized forms match exactly.

If no answer can be extracted from a generated continuation, the prefix is assigned the \noguess label and treated as non-equivalent. The verifier always compares $\hat{A}_i$ against the model's own final answer $\hat{A}_n$, rather than the benchmark ground-truth label, therefore detecting final answer confidence convergence, not task correctness.

\section{First token collisions between semantically different guesses}
\label{app:collision}

A first-token collision occurs at prefix $X_i$ when the forced target token $a_0$ (the first token of $\hat{A}_n$) is also the first token of a semantically different answer (i.e. $a_0 = a_n$ but $\hat{A}_i \not\equiv \hat{A}_n$), see example in~\Cref{tab:collision-example}. In this scenario, that we will be referring to as cross-class collisions, $p_i = P(a_0 \mid X_i)$ overestimates the model's commitment to $\hat{A}_n$: the model may be confidently generating a distinct answer whose surface form happens to share a leading token with $\hat{A}_n$.

For any prefix $X_i$ at which a cross-class collision is detected, the step confidence $p_i$ is unreliable as a proxy for commitment to $\hat{A}_n$, and the prefix is effectively excluded from the semantic attribution signal. Table~\ref{tab:dataset-collisions} reports collision rates across benchmarks.

\begin{table}[ht]
\centering
\small
\begin{tabular}{ll}
\toprule
\textbf{Full-CoT answer} $\hat{A}_n$& \texttt{49} \\
\textbf{First token} $a_0$ & \texttt{4} \\
\midrule
\textbf{Candidates at prefix $X_i$} & \\
\quad \textit{wrong-guess} & \texttt{420}, \texttt{406} \\
\quad \textit{no-guess} & \texttt{40}, \texttt{44} \\
\quad \textit{final-equivalent} & \texttt{49} \\
\bottomrule
\end{tabular}
\caption{\textbf{Example of a cross-class collision} on a trace (in detail: q16 of AIME2025, \texttt{Qwen3-14B}). The token \texttt{4} is shared by answers of different semantic classes, making $p_i$ an unreliable commitment signal.}
\label{tab:collision-example}
\end{table}

It is important to note that the collision rate is task-dependent. For instance, MATH-500, which involves a significant amount of \LaTeX-heavy responses, has a much higher collision rate compared to AIME in most of the cases, where all responses are numerical, or even rarer, GPQA-Diamond, which is a multiple-choice format and thus responses are predetermined from the outset.

\begin{table*}[ht]
\centering
\small
\begin{tabular}{lrrrrrr}
\toprule
\textbf{Benchmark} & \multicolumn{2}{c}{\texttt{gpt-oss-20b}} & \multicolumn{2}{c}{\texttt{gemma-4-26B-A4B-it}} & \multicolumn{2}{c}{\texttt{Qwen3-14B}} \\
\cmidrule(lr){2-3} \cmidrule(lr){4-5} \cmidrule(lr){6-7}
 & \textbf{Traces} & \textbf{Collision rate} & \textbf{Traces} & \textbf{Collision rate} & \textbf{Traces} & \textbf{Collision rate} \\
\midrule
MATH-500     & 7883 & 10.85\% & 2855 & 11.91\% & 2852 & 11.68\% \\
AIME-2025    &  367 &  0.82\% &   90 & 56.67\% &  124 & 12.10\% \\
ZebraLogic   &  335 &  2.99\% &  273 &  0.00\% &  300 &  4.00\% \\
GPQA-Diamond & 1432 &  0.28\% & 1004 &  0.00\% & 1210 &  1.24\% \\
\midrule
W. Average   &   -- &  8.71\% &   -- &  9.26\% &   -- &  8.36\% \\
\bottomrule
\end{tabular}
\caption{First-token cross-collision rates across benchmarks.}
\label{tab:dataset-collisions}
\end{table*}
The anomalously high collision rate observed for \texttt{gemma-4-26B-A4B-it} on AIME-2025 (56.67\%) is not a measurement artifact, but reflects a genuine property of the model's reasoning style on competition mathematics: \texttt{gemma-4-26B-A4B-it} frequently produces intermediate numerical guesses (e.g., \texttt{21}) that share a leading token with the final answer (expressed as \texttt{21+56-7=70}), causing multiple semantically distinct candidates to collapse onto the same first token despite being semantically distinct quantities.

\section{Clue sensitivity (\texorpdfstring{$\tau$}{tau}) ablation}\label{app:tau}
To assess sensitivity to the threshold $\tau$ (as defined in Section~\ref{sec:tau}), we study the \midguesses emergence by varying $\alpha \in \{0.3, 0.4, 0.5, 0.6, 0.7\}$: 
$$
\tau = p_0 + \alpha \cdot (p_n - p_0)
$$
and keeping all other hyper-parameters fixed.

In order to discriminate when an early-exit prediction is a \midguess or \textit{autoregressive noise}, we use $\tau$ as a sensitivity parameter to distinguish between \noguess and \midguess. A lower $\tau$ is more permissive: sentences carrying modest confidence gains are still credited as clues, increasing recall of intermediate guesses at the cost of potentially including low-confidence outputs. On the opposite, a higher $\tau$ is more conservative: only sentences that have already accumulated a large fraction of the full-CoT confidence gain are labelled as clues, trading recall for precision.

\begin{figure*}[t]
    \centering
    \includegraphics[width=0.85\textwidth]{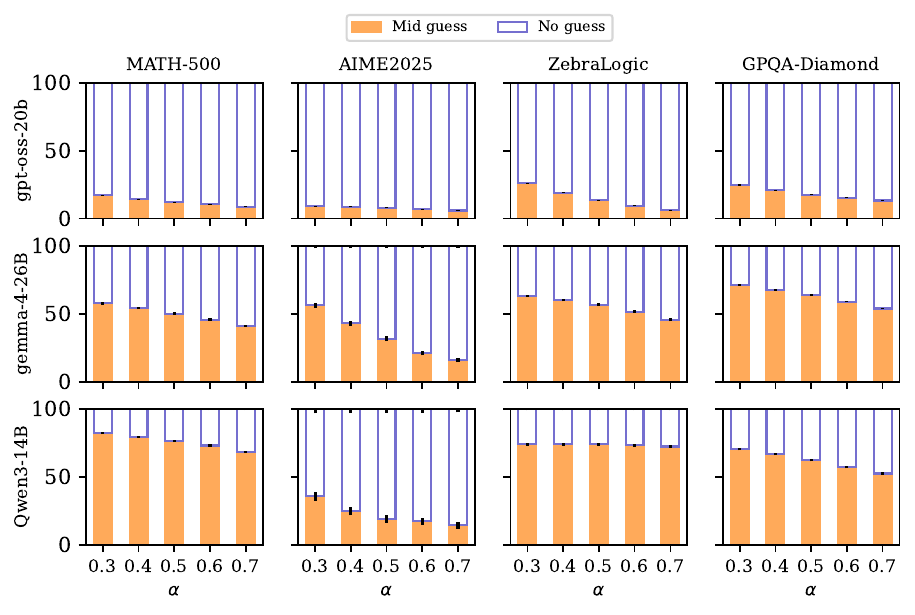}
    \caption{
        \textbf{\Midguess fraction as a function of $\tau$ across models and benchmarks.} Each panel shows the fraction of sentence spans classified as \midguess (orange) vs.\ \noguess (grey) for a given model (row) and benchmark (column), as $\tau$ varies in $\{0.3, 0.4, 0.5, 0.6, 0.7\}$.
        \Finalguesses are excluded from the denominator.
        \texttt{gpt-oss-20b} shows markedly fewer \midguesses on \textsc{AIME2025} compared to the other models, suggesting a more direct commitment pattern on hard mathematical reasoning.
    }
    \label{fig:tau-ablation}
\end{figure*}

Figure~\ref{fig:tau-ablation} reports the fraction of \midguess and \noguess spans across models and benchmarks for each value of $\tau$, thus excluding the \finalguesses from the total. As expected, increasing $\tau$ monotonically reduces the \midguess fraction in all settings, confirming that the parameter controls a smooth recall--precision trade-off. The effect is also dataset-dependent: on ZebraLogic, \midguesses dominate even at $\tau=0.7$, reflecting the high density of confident wrong intermediate answers in logical reasoning tasks. On \textsc{AIME2025}, \texttt{gpt-oss-20b} shows very few \midguesses across all $\tau$ values, suggesting it tends to commit directly to a final answer with little intermediate exploration; \texttt{gemma-4-26B-A4B-it} and \texttt{Qwen3-14B} exhibit instead a richer \midguess structure on the same benchmark.

\begin{figure*}[b]
    \centering
    \includegraphics[width=\textwidth, height=\textheight, keepaspectratio]{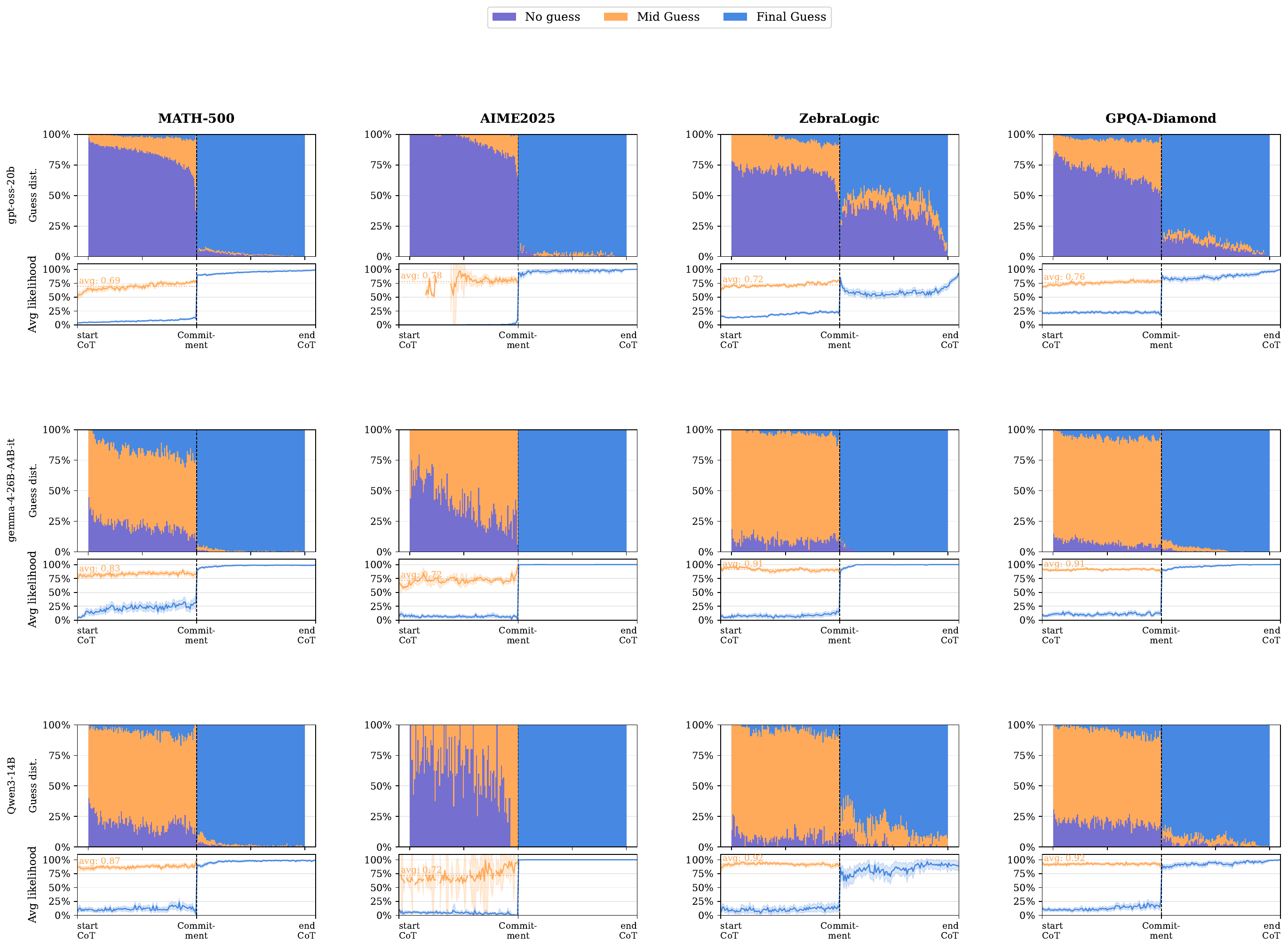}
    \caption{
        Guess distribution (top) and average token likelihood (bottom) across CoT positions, 
        relative to the commitment boundary, for all three models and four benchmarks.
        The $x$-axis is normalised so that first, second and third ticks correspond to start, commitment boundary $i^*$ and end of the CoT respectively.
        \textbf{Top panels:} stacked proportion of sentence spans labelled as \noguess, \midguess and \finalguess at each relative position.
        \textbf{Bottom panels:} token likelihood progresison of \midguess and \finalguess, with 95\% confidence intervals shaded.
        Across all models, \finalguess likelihood rises sharply at~$i^*$ and stabilises post-commitment, confirming that the boundary captures a genuine, abrupt shift in the model's output distribution.
        \midguess before $i^*$ remains consistently high for \texttt{gemma-4-26B-A4B-it} and \texttt{Qwen3-14B}, whereas for \texttt{gpt-oss-20b} it is more concentrated in the commitment proximity, consistent with a more decisive reasoning style.
    }
    \label{fig:midguess-grid}
\end{figure*}

While \texttt{gpt-oss-20b} concentrates almost all wrong-guess masses near zero, specifically on mathematical benchmarks, \texttt{gemma-4-26B-A4B-it} and \texttt{Qwen3-14B} exhibit broader, higher-confidence distributions. This implies that \texttt{gpt-oss-20b} candidate alternative solutions are spotted only for the transitory phase in which the model is seriously committed to the wrong answer, as stated in Section~\ref{sec:good-vs-useless} and shown in Figure~\ref{fig:midguess-logprobs}. For \texttt{gemma-4-26B-A4B-it} and \texttt{Qwen3-14B} the pattern is more noisy, as shown in Figure~\ref{fig:midguess-grid}.

\section{Post commitment boundary inflection points}
\label{app:post-boundary-examples}

Reasoning traces frequently contain linguistic markers associated with deliberation, namely words and phrases such as \textit{wait}, \textit{but}, \textit{check}, or \textit{let's}, that superficially signal reconsideration or uncertainty. We ask whether these \emph{verbal uncertainty signals} are systematically distributed relative to the commitment boundary $i^*$.

\begin{figure*}[ht]
    \centering
    \includegraphics[width=\linewidth]{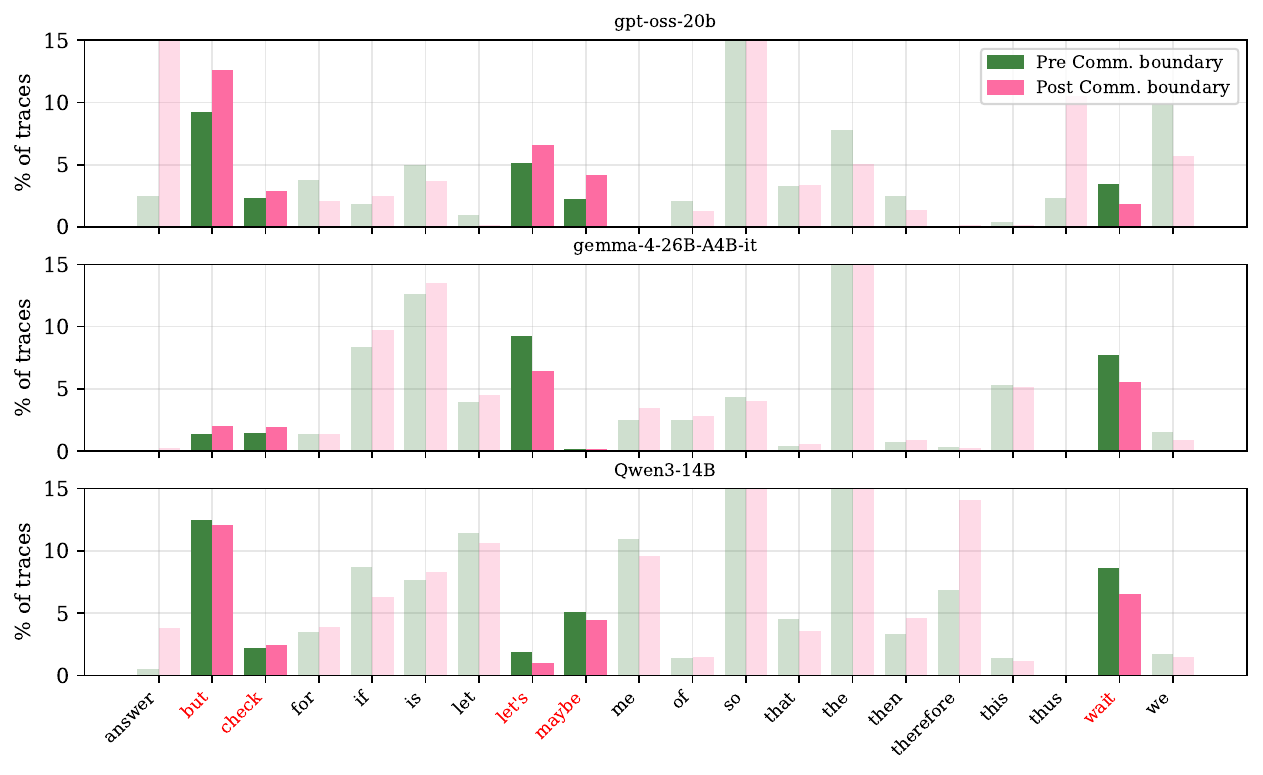}
    \caption{
        \textbf{Uncertainty-signalling language is equally frequent before and after the commitment boundary.} 
        We show the top-20 sentence-initial words across all models (e.g. roughly 12\% of Qwen3-14B sentences begin with "\textit{but}", at similar rates on both sides of $i^*$). Manually \textcolor{red!90!black}{highlighted} words can signal re-verification behaviour -- yet their frequency does not meaningfully change after the commitment boundary, confirming that such language is epiphenomenal: the model produces the same hedging and rechecking vocabulary even after our truncation experiments have causally established that no subsequent reasoning can alter its final answer. Results aggregated across MATH-500, AIME 2025, and GPQA-Diamond.
    }
    \label{fig:inflection-points}
\end{figure*}

For each trace, we compute the fraction of sentences whose first three words contain at least one such marker, separately for the pre- and post-boundary portions. 
We restrict our analysis to the first three words of each sentence, rather than scanning the full text, to target these markers in their role as deliberation signals: words like \textit{wait} or \textit{actually} are indicative of reconsideration when they open a new reasoning step, but lose this interpretation when they appear embedded in mathematical or syntactic contexts mid-sentence.
Figure~\ref{fig:inflection-points} reports these fractions across three models on MATH-500, AIME 2025, and GPQA-Diamond.

The distributions are largely similar on both sides of the boundary, with no marker showing a consistent directional shift across all three models. \textit{wait} and \textit{but} remain among the most frequent markers both before and after $i^*$; \textit{let's} shows a modest pre-boundary preference in \texttt{gemma-4-26B-A4B-it} and \texttt{gpt-oss-20b}, but not in \texttt{Qwen3-14B}. Overall, the presence of these markers alone is not a reliable indicator of whether the model is still actively updating its answer. The same surface signal can occur on either side of the commitment boundary.

\begin{figure*}[t]
    \centering
    \begin{subfigure}{\linewidth}
        \centering
        \includegraphics[width=\linewidth]{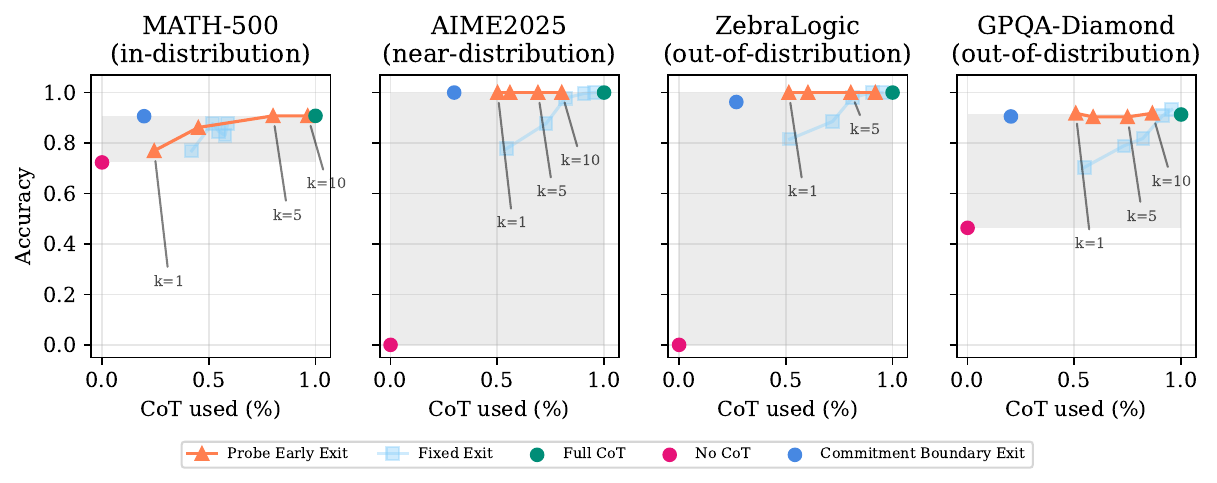}
        \caption{\textsc{gemma-4-26B-A4B-it}}
    \end{subfigure}

    \vspace{0.5cm}

    \begin{subfigure}{\linewidth}
        \centering
        \includegraphics[width=\linewidth]{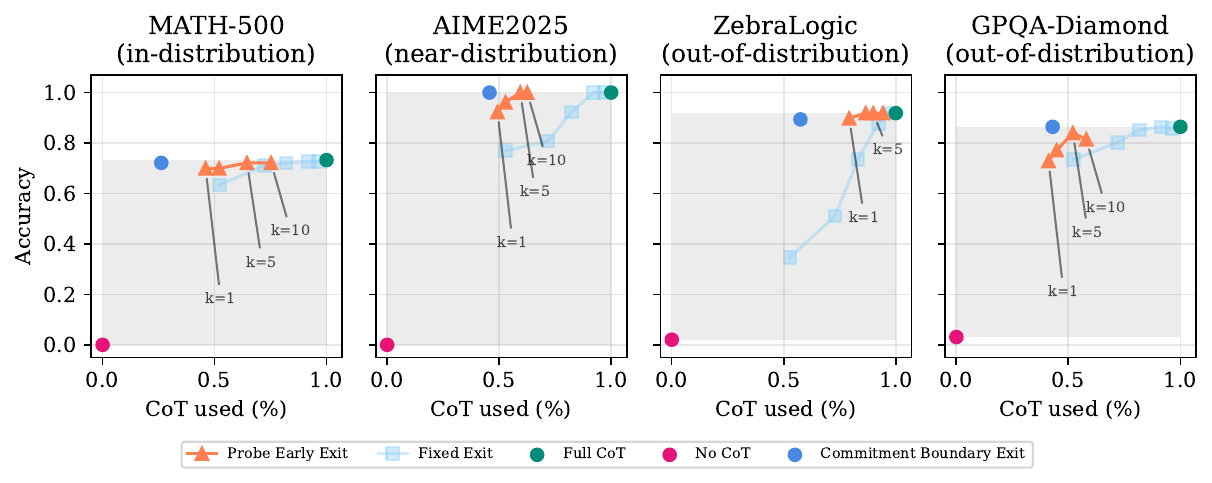}
        \caption{\textsc{Qwen3-14B}}
    \end{subfigure}

    \caption{Causal optimal early-exit accuracy versus CoT fraction across models for \textsc{gemma-4-26B-A4B-it} and \textsc{Qwen3-14B}.}
    \label{fig:combined_accuracy_vs_cot}
\end{figure*}

\section{Trace length is \texorpdfstring{\midguesses}{mid-guesses}-dependent}
\label{app:midguess}

Traces exhibiting at least one \midguess before $i^*$ tend to be longer than those that proceed directly to the final answer. To test whether this relationship holds monotonically with the number of \midguesses, we apply the Mann-Whitney $U$-test to compare the distribution of trace lengths between groups of traces binned by their number of \midguesses $n$. The test is non-parametric and does not assume normality, making it appropriate for the skewed token-length distributions observed in practice. We report results for both raw and $\tau$-gated \midguesses separately. Table~\ref{tab:mw-gpt} reports detailed results for \texttt{gpt-oss-20b} across all benchmarks.

On MATH-500, all consecutive comparisons are highly significant, with median trace length increasing monotonically from 252 tokens (zero \midguesses) to over 1000 tokens (more than six \midguesses). The same monotone pattern holds on GPQA-Diamond and ZebraLogic. On AIME-2025, where problems are substantially harder and the number of \midguesses is consistently higher, the relationship does not reach significance. We attribute this to high intrinsic variability in trace length across problems, where problem-specific structure dominates computation cost and obscures the \midguess signal.

\begin{table*}[t]
\centering
\small
\resizebox{\textwidth}{!}{%
\begin{tabular}{lrrllrr}
\toprule
& \multicolumn{3}{c}{\textbf{Standard \midguesses}} & \multicolumn{3}{c}{\textbf{$\tau$-gated \midguesses (Appendix~\ref{app:tau}))}} \\
\cmidrule(lr){2-4} \cmidrule(lr){5-7}
\textbf{Dataset} & \textbf{Comparison ($n$)} & \textbf{$p$-value} & \textbf{medians} & \textbf{Comparison ($n_1 / n_2$)} & \textbf{$p$-value} & \textbf{medians} \\
\midrule
\multirow{3}{*}{MATH-500}
& $0$ vs $1$ (493/781)          & $\mathbf{9.7 \times 10^{-31}}$ & 252 / 366   
& $0$ vs $1$ (2577/1125)    & $\mathbf{5.8 \times 10^{-37}}$ & 428 / 609  \\
& $1$ vs $[2,3]$ (781/1570)   & $\mathbf{1.2 \times 10^{-25}}$ & 366 / 488   
& $1$ vs $\geq2$ (1125/664)  & $\mathbf{2.4 \times 10^{-16}}$ & 609 / 904  \\
& $[2,3]$ vs $\geq4$ (1570/1522) & $\mathbf{1.8 \times 10^{-84}}$ & 488 / 893 &                               &                                &            \\
\midrule
\multirow{3}{*}{ZebraLogic}
& $[1,2]$ vs $3$ (59/80)      & $\mathbf{6.2 \times 10^{-13}}$ & 1573 / 3553 
& $1$ vs $2$ (21/63)        & $\mathbf{2.9 \times 10^{-5}}$  & 1126 / 2236 \\
& $3$ vs $>3$ (80/52)            & $\mathbf{1.9 \times 10^{-3}}$  & 3553 / 5910 & 
$2$ vs $3$ (63/76)        & $\mathbf{5.5 \times 10^{-8}}$  & 2236 / 4016 \\
&                                   &                                &             & 
$3$ vs $4$ (76/27)        & $\mathbf{4.1 \times 10^{-4}}$  & 4016 / 7034 \\
\midrule
\multirow{3}{*}{GPQA-Diamond}
& $0$ vs $1$ (21/250)           & $\mathbf{1.8 \times 10^{-4}}$  & 671 / 1165  
& $0$ vs $1$ (217/278)      & $\mathbf{1.3 \times 10^{-2}}$  & 1092 / 1395 \\
& $1$ vs $2$ (250/257)          & $\mathbf{3.8 \times 10^{-4}}$  & 1165 / 1485 & 
$1$ vs $\geq2$ (278/131)   & $\mathbf{1.7 \times 10^{-4}}$  & 1395 / 1800 \\
& $2$ vs $\geq3$ (257/98)        & $1.2 \times 10^{-1}$           & 1485 / 1654 &                               &                                &            \\
\bottomrule
\end{tabular}%
}\par
\caption{
    Mann--Whitney $U$-test results for \texttt{gpt-oss-20b} across all benchmarks.
    Standard and $\tau$-gated \midguesses (Appendix ~\ref{app:tau}) use dataset-specific bins chosen to balance statistical power and interpretability; gated bins are coarser where the $\tau$ threshold concentrates traces at low $n$. Significant $p$-values ($p < 0.05$) are in \textbf{bold}. Median trace lengths are in tokens.
}
\label{tab:mw-gpt}
\end{table*}

\newpage 

\section{Causal linear probe architecture}\label{app:probe}

The probe operates on sentence-level hidden states: at each sentence boundary $C_i$, it receives the hidden state $\mathbf{h}_i$ of the last token in that sentence and must predict the commitment label $y_i \in \{\noguess, \midguess, \finalguess\}$ using only information from sentences $1, \dots, i$ -- a causality constraint that makes the probe directly applicable during generation, without requiring the full trace to be completed first. Rather than pooling over the full trace prefix (which would conflate early and late reasoning steps and thus introduce a position bias), we restrict the attention to a local window of the $w$ most recent sentences.  
Formally, let $\mathbf{H} = [\mathbf{h}_1; \dots; \mathbf{h}_n] \in \mathbb{R}^{n \times d}$, and let $\tilde{\mathbf{H}} = \mathrm{LN}(\mathbf{H})$ be its layer-normalised version.
For each query position $i$, the probe pools causally over a window of the $w$ most recent sentences:

\begin{align}
    \alpha_{ij} &\propto \exp(\mathbf{w}_q^\top \tilde{\mathbf{h}}_j) \cdot \mathbb{1}[i-w < j \le i], \label{eq:attn} \\
    \mathbf{c}_i &= \textstyle\sum_{j} \alpha_{ij}\, W_v \tilde{\mathbf{h}}_j, \label{eq:context} \\
    \mathbf{z}_i &= \mathrm{ReLU}\big([\mathbf{c}_i \,\|\, W_\ell \tilde{\mathbf{h}}_i]\big), \label{eq:combine} \\
    f(\mathbf{h}_{\le i}) &= \mathrm{softmax}(W_o\, \mathbf{z}_i), \label{eq:probe}
\end{align}

where $W_q, W_v, W_\ell \in \mathbb{R}^{d \times h}$ and $W_o \in \mathbb{R}^{2h \times 3}$ are the only learned parameters.

The local term $W_\ell \tilde{\mathbf{h}}$ preserves the current sentence representation, while the causal pool $\mathbf{c}_i$ supplies prefix context constrained to the most recent $w$ sentences.

\section{Commitment Boundary Probe: Sweep Results}
\label{app:probe-sweep}

We describe the architecture and hyperparameter selection for the causal attention logistic probe
introduced in Section~\ref{sec:probe}.
The probe is trained to predict whether a given CoT step $i$ has a guess at all and if so, whether it is \textsc{mid-} or \textsc{final-} from the hidden state $\mathbf{h}_i$ at a fixed transformer layer. We evaluate three models: \texttt{gpt-oss-20b}, \texttt{gemma-4-26B-A4B-it} and \texttt{Qwen3-14B}.

\paragraph{Setup}
We sweep over three axes: (i) the transformer layer from which $\mathbf{h}_i$ is
extracted ($\{50\%, 80\%, 90\%, 100\%\}$ of total depth (exact layer correspondence in different model architectures is detailed in Table~\ref{tab:layer-correspondence}); (ii) the aggregation strategy over the hidden states of the sentence window preceding $C_i$: \texttt{last} uses only the hidden state of the final token of the window, while \texttt{avg} averages the hidden states of all tokens in the window; and (iii) the size of the context window fed to the probe ($\{8, 16, 32, 64, 128, 256\}$ tokens). All probes are trained on a held-out subset of MATH-500 traces and evaluated on a disjoint test set.

\begin{table}[H]
\centering
\small
\begin{tabular}{lcccc}
\toprule
\textbf{Model} & \textbf{50\%} & \textbf{80\%} & \textbf{90\%} & \textbf{100\%} \\
\midrule
\texttt{gpt-oss-20b}       & 12 & 19 & 22 & 23 \\
\texttt{gemma-4-26B-A4B-it} & 15 & 23 & 27 & 29 \\
\texttt{Qwen3-14B}         & 19 & 31 & 35 & 39 \\
\bottomrule
\end{tabular}
\caption{Layer index correspondence for each model architecture at the four depth percentiles used in the probe sweep.}
\label{tab:layer-correspondence}
\end{table}

We evaluate each configuration along four boundary-detection metrics at operating point $k=\{1,2,5, 10\}$ (i.e., the probe must fire on $k$ consecutive post-boundary steps before triggering):
\textbf{early-fire rate} (fraction of traces where the probe fires before $i^*$, lower is better),
\textbf{detect rate} (fraction of traces where the probe correctly identifies the post-boundary region, higher is better),
\textbf{miss rate} (fraction of traces where the probe never fires, lower is better), and
\textbf{saved fraction} (mean fraction of CoT tokens that would be skipped under early exit).
We also report \textbf{class-2 precision}, i.e. the precision of the probe on the post-boundary class.

\paragraph{Training}
We train on a stratified split of MATH-500 traces (50\% train / 10\% validation / 10\% test, the remaining 30\% held out for the early-exit experiments of Section~\ref{sec:early-exit}.
The split is at the question level, ensuring that no sentence from a held-out question appears at train time.
We optimise a class-weighted cross-entropy loss with weights inversely proportional to class frequency (clipped to a maximum ratio of $5$), Adam with learning rate $10^{-4}$ and weight decay $10^{-4}$, batch size $64$, gradient clipping at $1.0$, and dropout $0.1$.
Early stopping based on the validation macro-$F_1$ (patience $6$) selects the final checkpoint after $10$ epochs.
At inference we predict $\hat{y}_i = \arg\max f(\mathbf{h}_{\le i})$. 

Sensitivity of the $\tau$ threshold has already been discussed into Appendix~\ref{app:tau}.

\paragraph{Results}
Tables~\ref{tab:probe-sweep-gpt}, \ref{tab:probe-sweep-gemma} and \ref{tab:probe-sweep-qwen} report results for the top configurations of each tested model.

For \texttt{gpt-oss-20b} (Table~\ref{tab:probe-sweep-gpt}), the best overall configuration is \texttt{layer100\_last256} (layer 23, \texttt{last} aggregation, 256-token window), which achieves a $k{=}2$ early-fire rate of $6.87\%$, a detect rate of $93.13\%$, zero misses, and a class-2 precision of $91.59\%$, while saving $34.50\%$ of post-boundary tokens on average. 
Two clear trends emerge from the sweep.
First, later layers are more informative: layer 23 consistently outperforms layers 12 and 19 on early-fire and detect rate, confirming that commitment boundary information is encoded in deeper representations. Second, larger context windows improve probe quality up to 256 tokens, suggesting that the probe benefits from attending to a longer span of recent reasoning before making a prediction. \texttt{avg} aggregation was uniformly worse than \texttt{last}, indicating that the hidden state of the most recent token is the primary carrier of boundary information.

For \texttt{gemma-4-26B-A4B-it} (Table~\ref{tab:probe-sweep-gemma}), the best overall
configuration is \texttt{layer100\_last64} (layer 29, \texttt{last} aggregation, 64-token
window), which achieves an early-fire rate of $5\%$, a detect rate of $95\%$, zero misses,
and a class-2 precision of $99\%$.
The same depth trend observed for \texttt{gpt-oss-20b} holds: layer 29 dominates on
early-fire and detect rate, while shallower layers remain competitive on class-2 F1.
Notably, the context window exhibits an inverted pattern relative to \texttt{gpt-oss-20b}:
smaller windows (64 tokens) yield lower early-fire rates, suggesting that for this model
a tighter local context is sufficient and larger windows introduce noise.

For \texttt{Qwen3-14B} (Table~\ref{tab:probe-sweep-qwen}), the best overall configuration
is \texttt{layer90\_last64} (layer 35, \texttt{last} aggregation, 64-token window), which
achieves an early-fire rate of $5\%$, a detect rate of $95\%$, zero misses, and a class-2
precision of $97\%$.
The optimal depth shifts to $90\%$ rather than $100\%$, probably due to the higher number of layers compared to the other tested architectures (40 in place of the 24 of \texttt{gpt-oss-20b} and 30 of \texttt{gemma-4-26B-A4B-it}).
Window size again favours smaller contexts on early-fire, consistent with the pattern seen in \texttt{gemma-4-26B-A4B-it}. Across all three models, almost-zero miss rates are achieved consistently, confirming that the probe rarely suppresses a true boundary detection regardless of configuration. For all the tested models and configurations, \texttt{avg} aggregation is worse than \texttt{last}.

\begin{table*}[b]
\centering
\resizebox{\columnwidth}{!}{%
\begin{tabular}{lrcrccccccc}
\toprule
& & & & \multicolumn{4}{c}{\textbf{Boundary}} & \multicolumn{3}{c}{\textbf{Class-2}} \\
\cmidrule(lr){5-8} \cmidrule(lr){9-11}
\textbf{Config} & \textbf{Layer} & \textbf{Agg} & \textbf{Window} 
    & \textbf{Early} $\downarrow$ & \textbf{Detect} $\uparrow$ & \textbf{Miss} $\downarrow$ & \textbf{Saved} $\uparrow$
    & \textbf{P} $\uparrow$ & \textbf{R} $\uparrow$ & \textbf{F1} $\uparrow$ \\
\midrule
\texttt{layer100\_last512} & 23 & last & 512 & \textbf{0.06} & \textbf{0.93} & \textbf{0.00} & 0.34 & 0.91 & \textbf{0.83} & \textbf{0.87} \\
\textbf{\texttt{layer100\_last256}} & 23 & last & 256 & \textbf{0.06} & \textbf{0.93} & \textbf{0.00} & 0.35 & \textbf{0.92} & \textbf{0.83} & \textbf{0.87} \\
\texttt{layer100\_last128} & 23 & last & 128 & 0.12 & 0.89 & \textbf{0.00} & 0.36 & 0.87 & \textbf{0.83} & 0.85 \\
\texttt{layer100\_last64}  & 23 & last & 64  & 0.07 & 0.93 & \textbf{0.00} & 0.34 & 0.89 & \textbf{0.83} & 0.86 \\
\texttt{layer100\_last32}  & 23 & last & 32  & 0.08 & 0.92 & \textbf{0.00} & 0.35 & 0.89 & 0.81 & 0.85 \\
\texttt{layer100\_last16}  & 23 & last & 16  & 0.10 & 0.90 & \textbf{0.00} & 0.36 & 0.88 & 0.79 & 0.83 \\
\texttt{layer100\_last8}   & 23 & last & 8   & 0.15 & 0.85 & \textbf{0.00} & \textbf{0.38} & 0.86 & 0.77 & 0.82 \\
\midrule
\texttt{layer100\_avg64}   & 23 & avg  & 64  & 0.08 & 0.91 & 0.01 & 0.32 & 0.87 & 0.78 & 0.83 \\
\texttt{layer100\_avg16}   & 23 & avg  & 16  & 0.10 & 0.87 & 0.03 & 0.33 & 0.85 & 0.73 & 0.79 \\
\midrule
\texttt{layer90\_last256}  & 22 & last & 256 & 0.11 & 0.89 & \textbf{0.00} & 0.35 & 0.90 & 0.82 & 0.86 \\
\texttt{layer90\_last128}  & 22 & last & 128 & 0.11 & 0.89 & \textbf{0.00} & 0.35 & 0.86 & 0.82 & 0.84 \\
\midrule
\texttt{layer80\_last256}  & 19 & last & 256 & 0.10 & 0.90 & \textbf{0.00} & 0.35 & \textbf{0.92} & \textbf{0.83} & \textbf{0.87} \\
\texttt{layer80\_last128}  & 19 & last & 128 & 0.09 & 0.91 & \textbf{0.00} & 0.35 & 0.89 & \textbf{0.83} & 0.86 \\
\texttt{layer80\_last64}   & 19 & last & 64  & 0.08 & 0.92 & \textbf{0.00} & 0.34 & 0.89 & \textbf{0.83} & 0.86 \\
\midrule
\texttt{layer50\_last16}   & 12 & last & 16  & 0.11 & 0.89 & \textbf{0.00} & 0.37 & 0.83 & \textbf{0.83} & 0.84 \\
\bottomrule
\end{tabular}
}
\caption{
    \textbf{Probe sweep results on MATH-500 (\texttt{gpt-oss-20b}).}
    All boundary metrics are reported at operating point $k{=}2$.
    Early-fire and miss rates are lower-is-better; all others are higher-is-better.
    The best configuration per column is \textbf{bold}.
}
\label{tab:probe-sweep-gpt}
\end{table*}

\begin{table*}[t]
\centering
\resizebox{\columnwidth}{!}{%
\begin{tabular}{lrcrccccccc}
\toprule
& & & & \multicolumn{4}{c}{\textbf{Boundary}} & \multicolumn{3}{c}{\textbf{Class-2}} \\
\cmidrule(lr){5-8} \cmidrule(lr){9-11}
\textbf{Config} & \textbf{Layer} & \textbf{Agg} & \textbf{Window}
    & \textbf{Early} $\downarrow$ & \textbf{Detect} $\uparrow$ & \textbf{Miss} $\downarrow$ & \textbf{Saved} $\uparrow$
    & \textbf{P} $\uparrow$ & \textbf{R} $\uparrow$ & \textbf{F1} $\uparrow$ \\
\midrule
\texttt{layer100\_last256}       & 29 & last &  256 & 0.23 & \textbf{0.77} & \textbf{0.00} & 0.77 & 0.97 & \textbf{0.93} & \textbf{0.95} \\
\texttt{layer100\_last128}       & 29 & last &  128 & 0.08 & 0.92 & \textbf{0.00} & 0.67 & \textbf{0.99} & 0.84 & 0.91 \\
\textbf{\texttt{layer100\_last64}  }     & 29 & last &   64 & \textbf{0.05} & \textbf{0.95} & \textbf{0.00} & 0.66 & \textbf{0.99} & 0.85 & 0.91 \\
\texttt{layer100\_last32}       & 29 & last &  32 & 0.07 & 0.93 & \textbf{0.00} & 0.66 & \textbf{0.99} & 0.82 & 0.89 \\
\texttt{layer100\_last16}       & 29 & last &  16 & 0.09 & 0.90 & 0.01 & 0.66 & \textbf{0.99} & 0.84 & 0.91 \\
\texttt{layer100\_last8}       & 29 & last &  8 & 0.14 & 0.86 & \textbf{0.00} & 0.70 & 0.98 & 0.81 & 0.89 \\
\midrule
\texttt{layer100\_avg64}       & 29 & avg &  64 & 0.16 & 0.81 & 0.04 & 0.66 & 0.97 & 0.83 & 0.90 \\
\texttt{layer100\_avg16}       & 29 & avg &  16 & 0.19 & 0.79 & 0.07 & 0.68 & 0.95 & 0.79 & 0.87 \\
\midrule
\texttt{layer90\_last256}        & 27 & last &  256 & 0.10 & 0.90 & \textbf{0.00} & 0.70 & 0.98 & 0.89 & 0.93 \\
\texttt{layer90\_last128}        & 27 & last &  128 & 0.06 & 0.94 & \textbf{0.00} & 0.68 & \textbf{0.99} & 0.88 & 0.93 \\
\texttt{layer90\_last64}         & 27 & last &   64 & 0.08 & 0.92 & \textbf{0.00} & 0.67 & \textbf{0.99} & 0.84 & 0.91 \\
\midrule
\texttt{layer80\_last256}        & 23 & last &  256 & 0.12 & 0.88 & \textbf{0.00} & 0.72 & 0.98 & 0.87 & 0.92 \\
\texttt{layer80\_last128}        & 23 & last &  128 & 0.10 & 0.90 & \textbf{0.00} & 0.71 & \textbf{0.99} & 0.90 & 0.94 \\
\texttt{layer80\_last64}         & 23 & last &   64 & 0.08 & 0.92 & \textbf{0.00} & 0.70 & \textbf{0.99} & 0.91 & 0.94 \\
\midrule
\texttt{layer50\_last256}        & 15 & last &  256 & 0.17 & 0.83 & \textbf{0.00} & 0.73 & 0.98 & 0.90 & 0.94 \\
\texttt{layer50\_last128}        & 15 & last &  128 & 0.11 & 0.89 & \textbf{0.00} & 0.73 & 0.98 & 0.88 & 0.93 \\
\texttt{layer50\_last64}         & 15 & last &   64 & 0.09 & 0.91 & \textbf{0.00} & 0.72 & \textbf{0.99} & 0.92 & \textbf{0.95} \\
\bottomrule
\end{tabular}}
\caption{
    \textbf{Probe sweep results on MATH-500 (\texttt{gemma-4-26B-A4B-it}).}
    All boundary metrics are reported at operating point $k{=}2$.
    Early-fire and miss rates are lower-is-better; all others are higher-is-better.
    The best configuration per column is \textbf{bold}.
}
\label{tab:probe-sweep-gemma}
\end{table*}

\begin{table*}[t]
\centering
\resizebox{\columnwidth}{!}{%
\begin{tabular}{lrcrccccccc}
\toprule
& & & & \multicolumn{4}{c}{\textbf{Boundary}} & \multicolumn{3}{c}{\textbf{Class-2}} \\
\cmidrule(lr){5-8} \cmidrule(lr){9-11}
\textbf{Config} & \textbf{Layer} & \textbf{Agg} & \textbf{Window}
    & \textbf{Early} $\downarrow$ & \textbf{Detect} $\uparrow$ & \textbf{Miss} $\downarrow$ & \textbf{Saved} $\uparrow$
    & \textbf{P} $\uparrow$ & \textbf{R} $\uparrow$ & \textbf{F1} $\uparrow$ \\
\midrule
\texttt{layer100\_last256}  & 39 & last & 256 & 0.28 & 0.72 & \textbf{0.00} & 0.68 & 0.92 & 0.93 & 0.93 \\
\texttt{layer100\_last64}   & 39 & last &  64 & 0.19 & 0.81 & \textbf{0.00} & 0.68 & 0.95 & 0.92 & 0.93 \\
\texttt{layer100\_last16}   & 39 & last &  16 & 0.24 & 0.76 & \textbf{0.00} & \textbf{0.69} & 0.93 & 0.89 & 0.91 \\
\midrule
\texttt{layer100\_avg64}    & 39 & avg  &  64 & 0.19 & 0.81 & \textbf{0.00} & 0.64 & 0.95 & 0.85 & 0.90 \\
\texttt{layer100\_avg16}    & 39 & avg  &  16 & 0.15 & 0.85 & \textbf{0.00} & 0.65 & 0.94 & 0.87 & 0.90 \\
\midrule
\texttt{layer90\_last256}   & 35 & last & 256 & 0.22 & 0.78 & \textbf{0.00} & 0.68 & 0.94 & 0.92 & 0.93 \\
\texttt{layer90\_last128}   & 35 & last & 128 & 0.20 & 0.80 & \textbf{0.00} & 0.68 & 0.94 & 0.93 & 0.94 \\
\textbf{\texttt{layer90\_last64}}    & 35 & last &  64 & \textbf{0.05} & \textbf{0.95} & \textbf{0.00} & 0.66 & \textbf{0.97} & 0.91 & 0.94 \\
\texttt{layer90\_last32}    & 35 & last &  32 & 0.16 & 0.84 & \textbf{0.00} & 0.65 & 0.93 & 0.89 & 0.91 \\
\texttt{layer90\_last16}    & 35 & last &  16 & 0.12 & 0.88 & \textbf{0.00} & 0.67 & 0.95 & 0.90 & 0.93 \\
\texttt{layer90\_last8}     & 35 & last &   8 & 0.24 & 0.76 & \textbf{0.00} & 0.68 & 0.93 & 0.87 & 0.90 \\
\midrule
\texttt{layer80\_last256}   & 31 & last & 256 & 0.18 & 0.82 & \textbf{0.00} & 0.67 & 0.95 & 0.91 & 0.93 \\
\texttt{layer80\_last128}   & 31 & last & 128 & 0.18 & 0.82 & \textbf{0.00} & 0.67 & 0.94 & 0.92 & 0.93 \\
\texttt{layer80\_last64}    & 31 & last &  64 & 0.08 & 0.92 & \textbf{0.00} & 0.65 & 0.96 & 0.91 & 0.93 \\
\midrule
\texttt{layer50\_last256}   & 19 & last & 256 & 0.11 & 0.89 & \textbf{0.00} & 0.66 & 0.96 & 0.92 & 0.94 \\
\texttt{layer50\_last128}   & 19 & last & 128 & 0.09 & 0.91 & \textbf{0.00} & 0.66 & \textbf{0.97} & 0.93 & \textbf{0.95} \\
\texttt{layer50\_last64}    & 19 & last &  64 & 0.19 & 0.81 & \textbf{0.00} & 0.67 & 0.94 & \textbf{0.94} & 0.94 \\
\bottomrule
\end{tabular}}
\caption{
    \textbf{Probe sweep results on MATH-500 (\texttt{Qwen3-14B}).}
    All boundary metrics are reported at operating point $k{=}2$.
    Early-fire and miss rates are lower-is-better; all others are higher-is-better.
    The best configuration per column is \textbf{bold}.
}
\label{tab:probe-sweep-qwen}
\end{table*}

\paragraph{Selected configuration}
The best probe according to the sweep procedure is adopted as our default probe throughout the paper for early-exit experiments detailed into Section~\ref{sec:early-exit}. \texttt{gemma-4-26B-A4B-it} and \texttt{Qwen3-14B} early-exit performance plots can be found in Figure~\ref{fig:combined_accuracy_vs_cot}.

These results confirm the claim in Section~\ref{sec:probe} that the commitment
boundary is linearly decodable from the model's hidden states with high reliability.

\section{Model generation arguments and reproducibility details}
All reasoning traces were generated using vLLM with \texttt{max\_model\_len=16\,384} tokens and we used tokens id instead of plain text to preserve end-of-thinking markers, an information used at attribution time to assess whether or not the trace has been truncated and thus the model has not reached its natural reasoning ending. Prompts were formatted via each model's chat template using \texttt{transformers}'s \texttt{AutoTokenizer.apply\_chat\_template}. Reasoning effort was kept to medium where available. For each question, 16 traces were sampled for \texttt{gpt-oss-20b} and 8 traces for all other models; for the latter ones, attribution was computed on up to 4 valid traces per question (i.e.\ traces where the full CoT strictly outperforms the no-CoT baseline and do not cause Out Of Memory errors).

As per generation parameters we used \texttt{temperature=0.7}, \texttt{top\_p=0.9} across all our model suite. We additionally employ \texttt{NNsight}~\citep{fiottokaufman2024nnsightndifdemocratizingaccess}.

The distinct pair of start/end markers to delimit its chain-of-thought reasoning section is model-dependent, and we show them as follows:

\begin{center}

\footnotesize
\begin{tabular}{lp{4.7cm}}
\toprule
\textbf{Model} & \texttt{[BOT]} (start) and \texttt{[EOT]} (end) tokens \\
\midrule
\texttt{gpt-oss-20b}          & \texttt{<|channel|>analysis<|message|>} \textbf{/} \texttt{<|channel|>final<|message|>} \\
\texttt{Qwen3-14B}            & \texttt{<think>} \textbf{/} \texttt{</think>} \\
\texttt{gemma-4-26B}   & \texttt{<|channel>thought} \textbf{/} \texttt{<channel|>} \\
\bottomrule
\end{tabular}
\end{center}

\section{Use of AI}

We used AI-based coding agents to assist with implementing experimental infrastructure and analysis scripts. All generated code was manually reviewed, validated, and, where necessary, modified by the authors, who take full responsibility for the final codebase and all reported results. Experimental designs, evaluation protocols, and scientific conclusions were conceived and verified by the authors; no experimental results or findings were autonomously generated by AI systems. AI-assisted writing tools were also used during manuscript preparation to improve clarity and presentation.

\end{document}